\documentclass[runningheads]{llncs}

\usepackage[mobile]{eccv}

\usepackage{eccvabbrv}
\usepackage{bbding}
\usepackage{multirow}
\usepackage{pifont}
\usepackage{float}
\usepackage{graphicx}
\usepackage{booktabs}
\usepackage{xcolor}
\usepackage[export]{adjustbox}
\usepackage{colortbl}
\usepackage{url}
\usepackage[final]{changes}

\def\xnet{AdaIR\xspace}

\usepackage[accsupp]{axessibility}  % Improves PDF readability for those with disabilities.

\usepackage{orcidlink}

\begin{document}
\nolinenumbers
\title{\xnet: Adaptive All-in-One Image Restoration via Frequency Mining and Modulation} 

\titlerunning{\xnet}

\author{Yuning Cui\inst{1} \and
Syed Waqas Zamir\inst{2}\and
Salman Khan\inst{3,4}\and\\
Alois Knoll\inst{1}\and
Mubarak Shah\inst{5}\and
Fahad Shahbaz Khan\inst{3,6}}

\authorrunning{Y. Cui et al.}

\institute{$^{1}$ Technical University of Munich   ~~$^{2}$ Inception Institute of Artificial Intelligence\\
$^{3}$ Mohammed Bin Zayed University of AI  ~~$^{4}$ Australian National University\\
$^{5}$ University of Central Florida  ~~$^{6}$ Linköping University}

% ,\and
% Inception Institute of Artificial Intelligence \and
% Mohammed Bin Zayed University of AI
% \and
% Australian National University \and
% University of Central Florida\and
% Linköping University
\maketitle

\begin{abstract}
In the image acquisition process, various forms of degradation, including noise, blur, haze, and rain, are frequently introduced. These degradations typically arise from the inherent limitations of cameras or unfavorable ambient conditions. To recover clean images from their degraded versions, numerous specialized restoration methods have been developed, each targeting a specific type of degradation.
Recently, all-in-one algorithms have garnered significant attention by addressing different types of degradations within a \emph{single model} without requiring the prior information of the input degradation type.
However, these methods purely operate in the spatial domain and do not delve into 
the distinct frequency variations inherent to different degradation types.
To address this gap, we propose an adaptive all-in-one image restoration network based on frequency mining and modulation. 
Our approach is motivated by the observation that different degradation types impact the image content on different frequency subbands, thereby requiring different treatments for each restoration task.
Specifically, we first mine low- and high-frequency information from the input features, guided by the adaptively decoupled spectra of the degraded image. 
The extracted features are then modulated by a bidirectional operator to facilitate interactions between different frequency components. 
Finally, the modulated features are merged into the original input for a progressively guided restoration. 
With this approach, the model achieves adaptive reconstruction by accentuating the informative frequency subbands according to different input degradations. 
Extensive experiments demonstrate that the proposed method, named \xnet, achieves state-of-the-art performance on different image restoration tasks, including image denoising, dehazing, deraining, motion deblurring, and low-light image enhancement. Our code and pre-trained models are available at \href{https://github.com/c-yn/AdaIR}{https://github.com/c-yn/AdaIR}.

\end{abstract}

\section{Introduction}

Image restoration is the task of generating a high-quality clean image by removing degradations (\eg, noise, haze, blur, rain) from the original input image. It serves as a vital component in numerous downstream applications across diverse domains, including image/video content creation, surveillance, medical imaging, and remote sensing. Given its inherently ill-posed nature, effective image restoration demands learning strong image priors from large-scale data. To this end, deep neural network-based image restoration approaches~\cite{ren2020single,dong2020multi,zamir2020cycleisp,ren2021adaptivedeamnet,zhang2017learning,tsai2022banet,nah2022clean,zhang2020deblurring} have emerged as preferable choices over the conventional handcrafted algorithms~\cite{he2010single,liu2020trident,dong2011image,timofte2013anchored,kim2010single,michaeli2013nonparametric,kopf2008deep}. Deep-learning methods learn image priors either implicitly from data~\cite{ren2021adaptivedeamnet,zhang2017learning,nah2022clean,zhang2020deblurring,ren2020single,dong2020multi,liu2020trident}, or explicitly by incorporating task-specific knowledge into the network architectures~\cite{tu2022maxim, restormer, wang2022uformer, Zamir_2021_CVPR_mprnet, zamir2020mirnet, chen2022simple}. Despite promising results on individual restoration tasks, these approaches are either not generalizable beyond the specific degradation types and levels which hinders their broader application, or require training separate copies of the same network on different degradation types, which is computationally expensive and tedious procedure, and maybe infeasible solution for deployment on resource-constraint edge-devices. Therefore, there is a need to develop an all-in-one image restoration method that can handle images with different degradation types, without requiring prior information regarding the corruption present in the input images.

\begin{figure}[t]
    \centering
    \begin{minipage}[l]{0.71\textwidth}
   \setlength\tabcolsep{0.3pt}
    \resizebox{\linewidth}{!}{
     \begin{tabular}{ccccc}
    \includegraphics[height=0.12\linewidth,width = 0.19\textwidth]{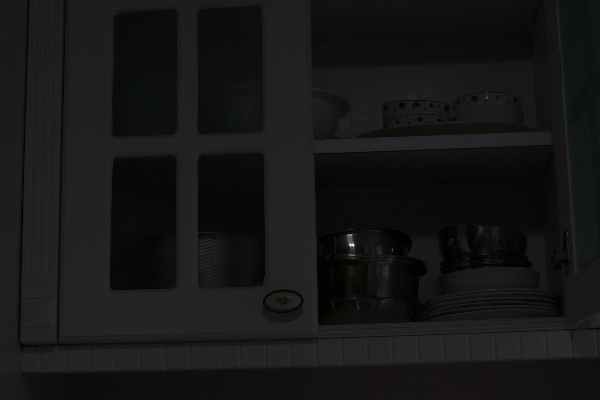}
    &\includegraphics[height=0.12\linewidth,width = 0.19\textwidth]{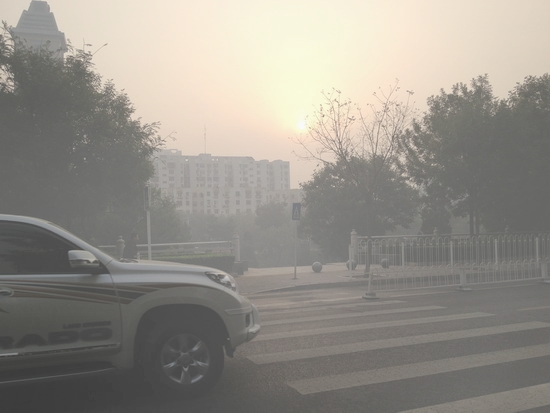}
    &\includegraphics[height=0.12\linewidth,width = 0.19\textwidth]{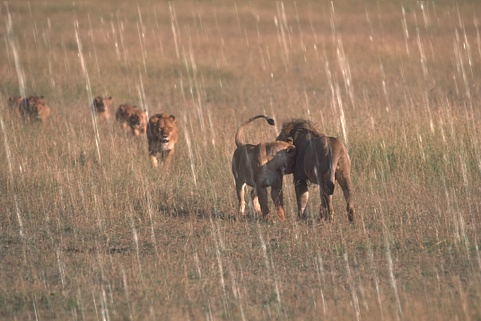}
    &\includegraphics[height=0.12\linewidth,width = 0.19\textwidth]{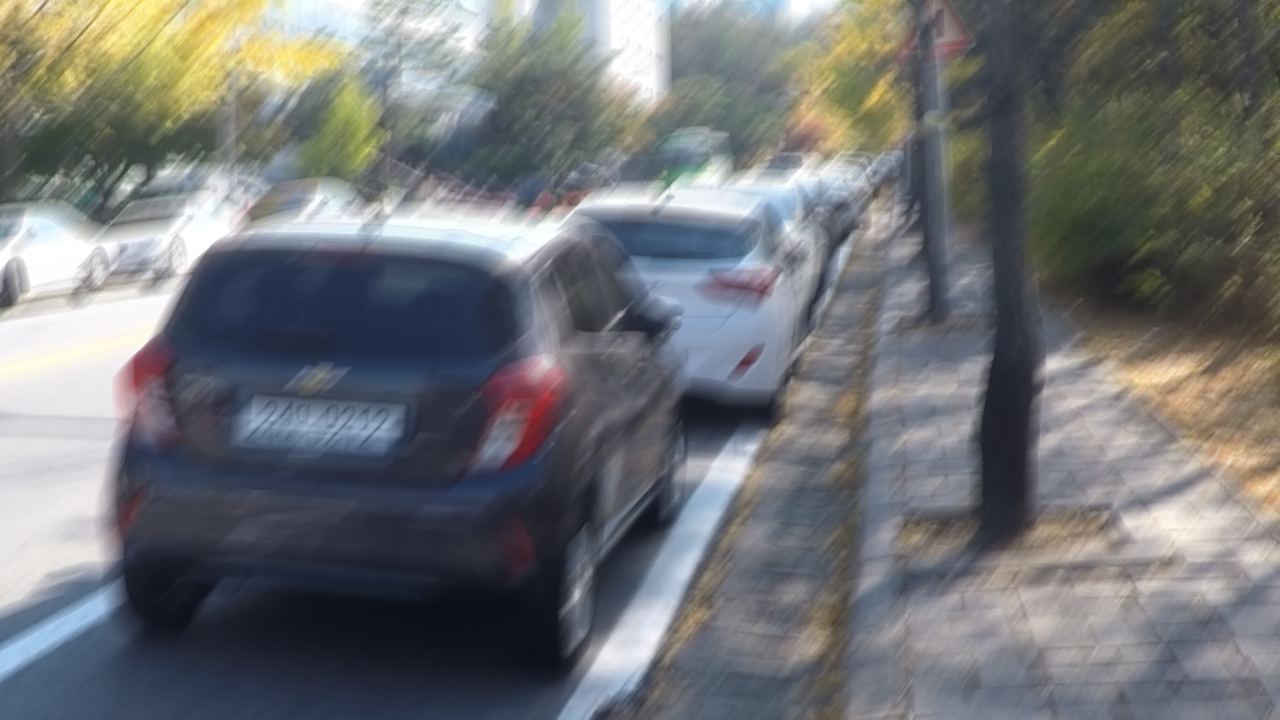}
    &\includegraphics[height=0.12\linewidth,width = 0.19\textwidth]{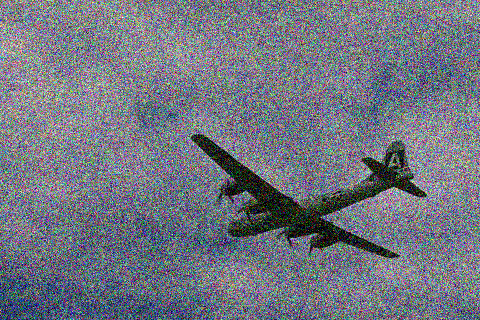} \\
    \includegraphics[height=0.12\linewidth,width = 0.19\textwidth]{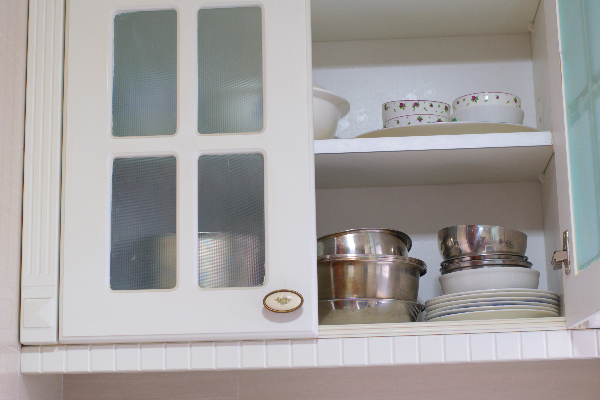}
    &\includegraphics[height=0.12\linewidth,width = 0.19\textwidth]{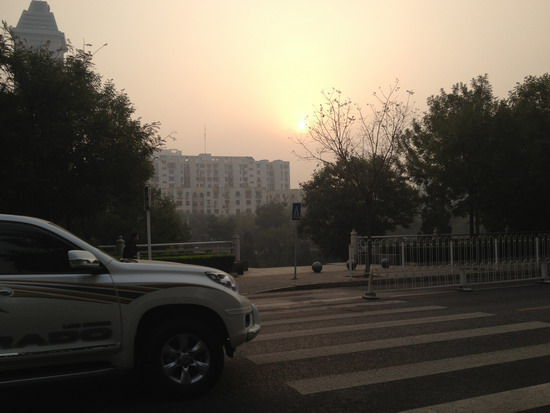}
    &\includegraphics[height=0.12\linewidth,width = 0.19\textwidth]{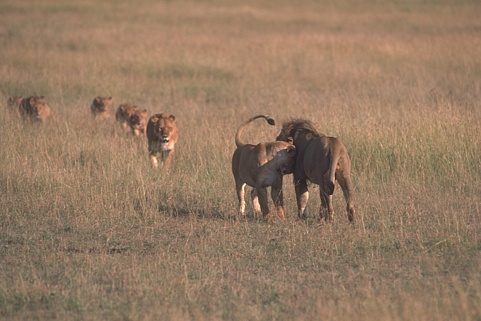}
    &\includegraphics[height=0.12\linewidth,width = 0.19\textwidth]{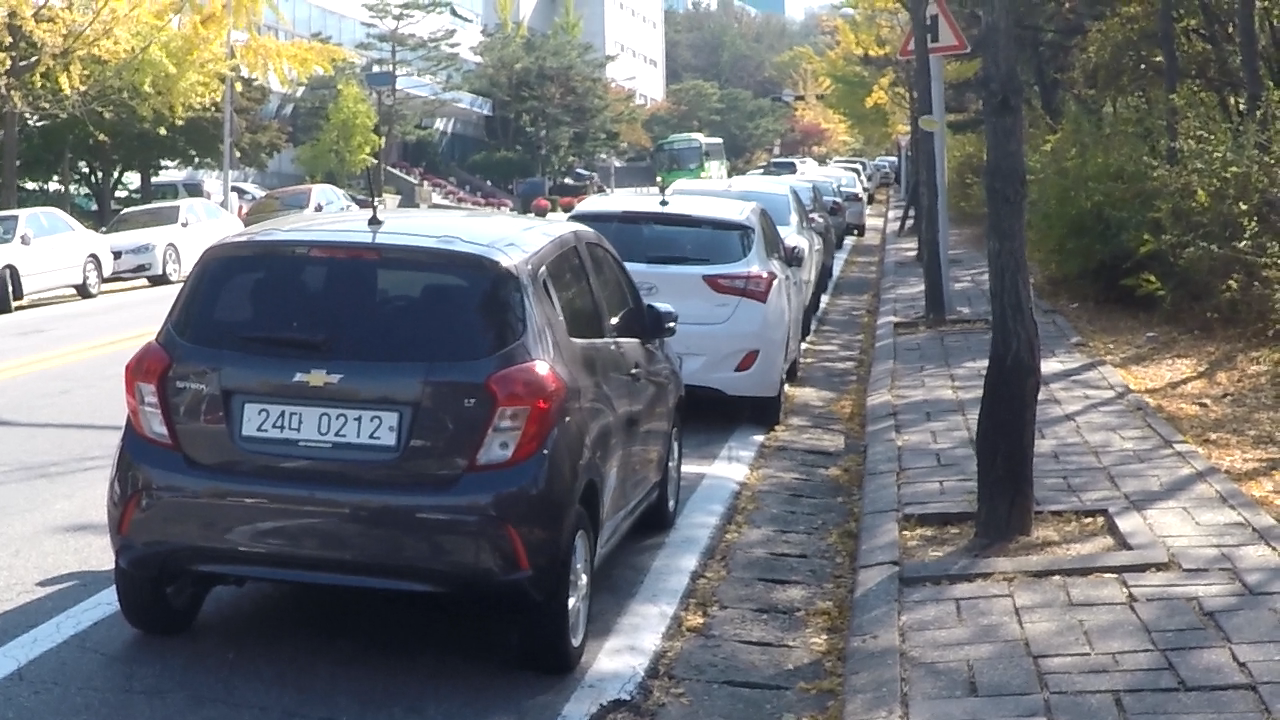}
    &\includegraphics[height=0.12\linewidth,width = 0.19\textwidth]{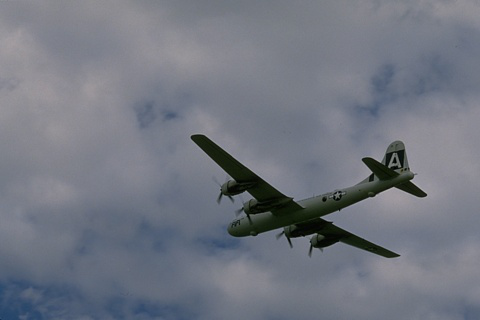} \\
    \includegraphics[height=0.12\linewidth,width = 0.19\textwidth]{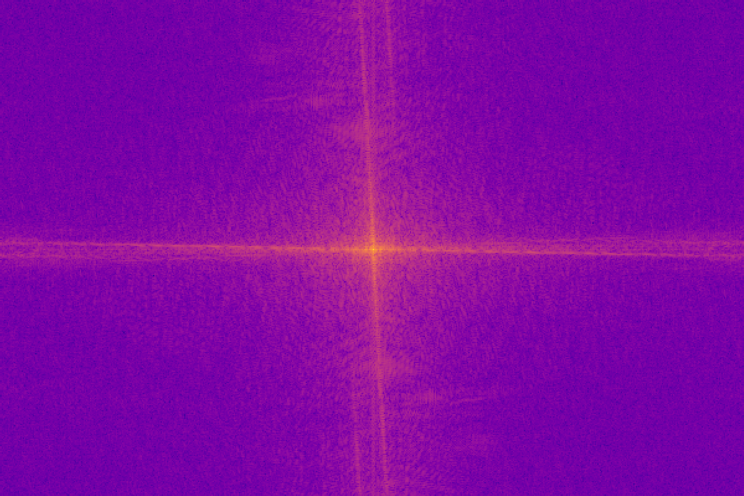}
    &\includegraphics[height=0.12\linewidth,width = 0.19\textwidth]{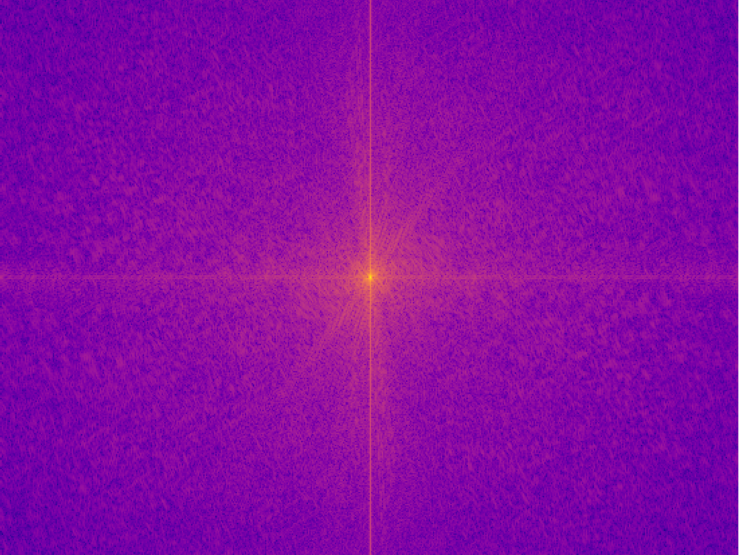}
    &\includegraphics[height=0.12\linewidth,width = 0.19\textwidth]{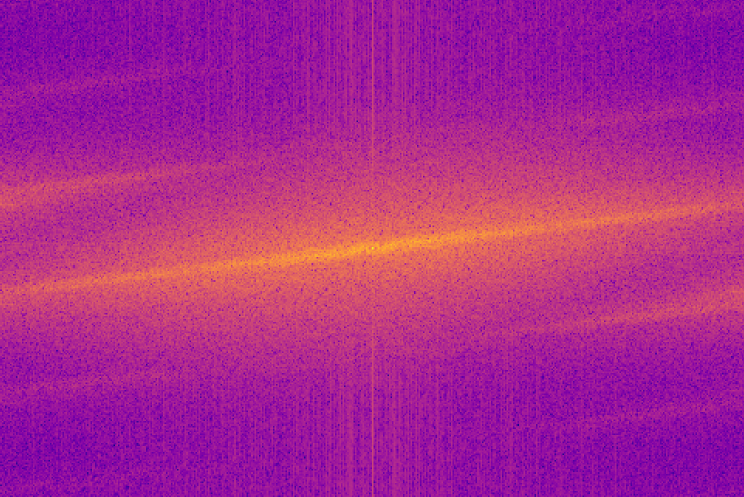}
    &\includegraphics[height=0.12\linewidth,width = 0.19\textwidth]{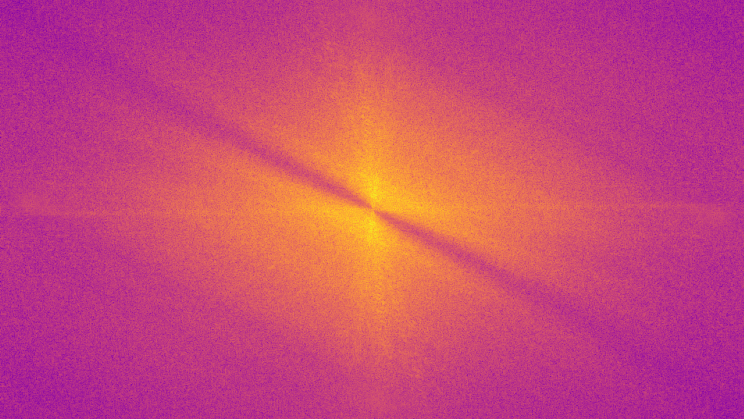}
    &\includegraphics[height=0.12\linewidth,width = 0.19\textwidth]{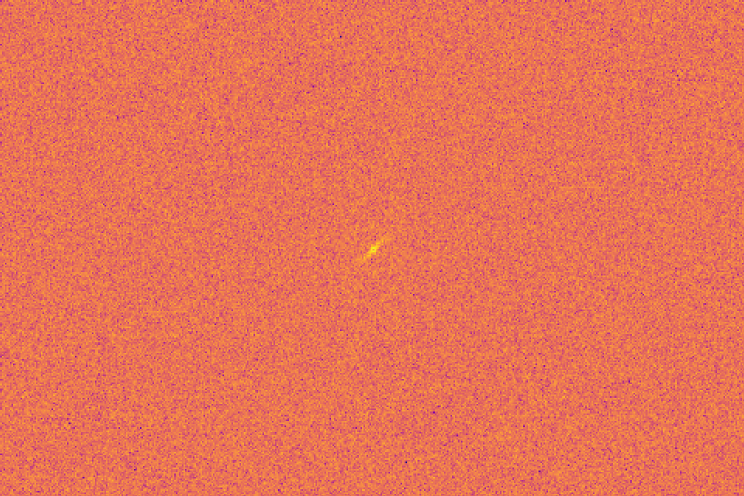}\\
    Low-Light & Dehazing & Deraining & Deblurring & Denoising \\

    \end{tabular}}
    \end{minipage}
    \begin{minipage}[c]{0.28\textwidth}
          \includegraphics[width = 1\textwidth]{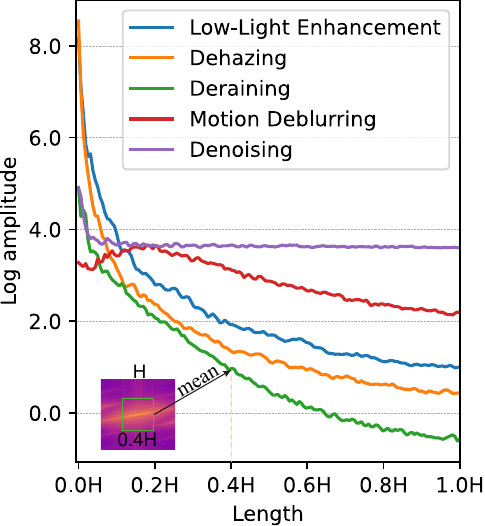}
    \end{minipage}
    \caption{\emph{Left}, from top to bottom: degraded images, ground-truth images, and the Fourier spectra of residual images obtained by subtracting the degraded images from the ground-truth images. The images are obtained from LOL-v1~\cite{lol}, SOTS~\cite{RESIDE}, Rain100L~\cite{RESIDE}, GoPro~\cite{gopro}, and BSD68~\cite{bsd68}, respectively.
    \emph{Right}, the sub-graph illustrates the mean values of Fourier spectra on the square of length shown on the x-axis, across five tasks. The spectra are all resized to $320\times 320$ for comparisons. As seen, different tasks pay different attention to different frequency subbands. For example, there are larger discrepancies in low frequency between degraded and target image pairs of the low-light image enhancement and dehazing datasets. In contrast, the frequency differences are generally evenly distributed for image denoising.}
    \label{fig:motivation}
    % \vspace{-0.6cm}
\end{figure}

Recently, an increasing number of attempts have been made~\cite{airnet, idr, promptir, ma2023prores} to address multiple degradations with a single model. These include employing a degradation-aware encoder in the restoration network learned via contrastive learning paradigm~\cite{airnet}; designing a two-stage framework IDR~\cite{idr}, where the first stage is dedicated to task-oriented knowledge collection based on underlying physics characteristics of degradation types, and the second stage is responsible for ingredients-oriented knowledge integration that progressively restores the image; or developing prompt-learning strategies~\cite{promptir, ma2023prores} inspired from their success in the natural language processing domain~\cite{brown2020language,shrivastava2023repository,lester2021power}. Nonetheless, all these approaches purely operate in the spatial domain, and do not consider frequency domain information. However, as illustrated in Fig.~\ref{fig:motivation}, we observe that different types of degradations may impact the image content on different frequency subbands. For instance, on the one hand, noisy and rainy images are contaminated with high-frequency content, while on the other hand, low-light and hazy images are dominated by low-frequency degraded content, thus indicating the need to treat each restoration task on its own merits.

\begin{figure}[t]
    \centering
        \centering
        \begin{tabular}{cccc}
         \includegraphics[height=0.2\linewidth,width = 0.26\textwidth]{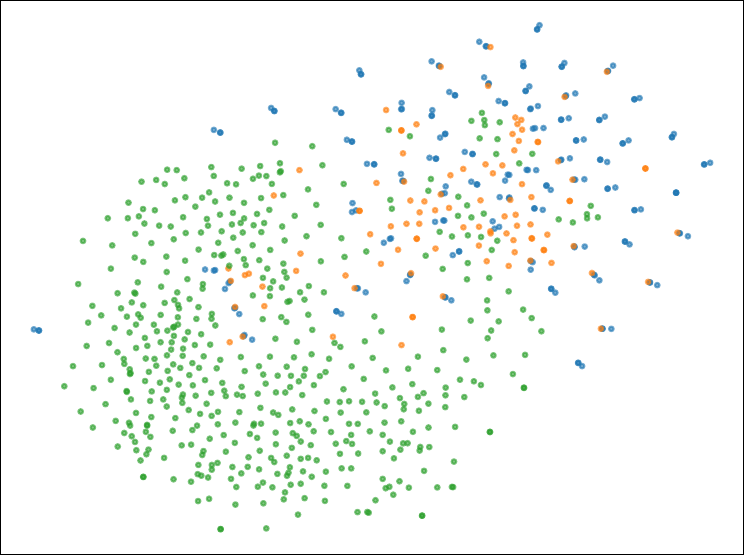}  
         &\includegraphics[height=0.2\linewidth,width = 0.26\textwidth]{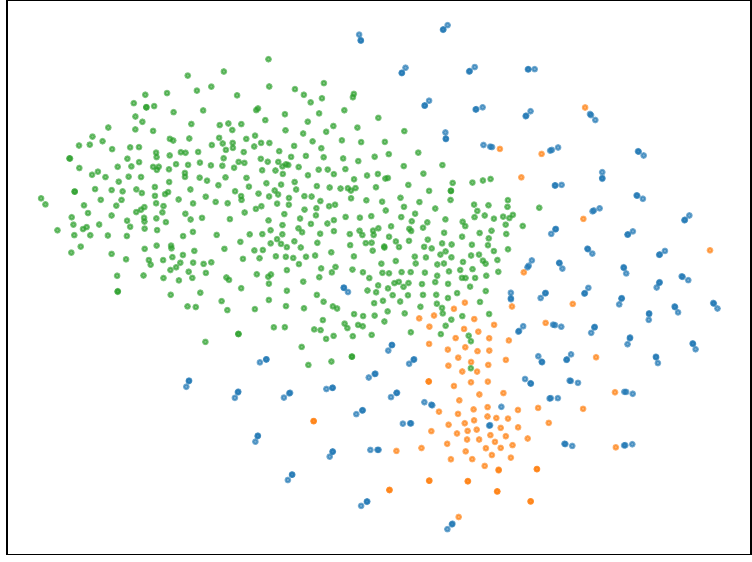}  
         &\includegraphics[height=0.2\linewidth,width = 0.26\textwidth]{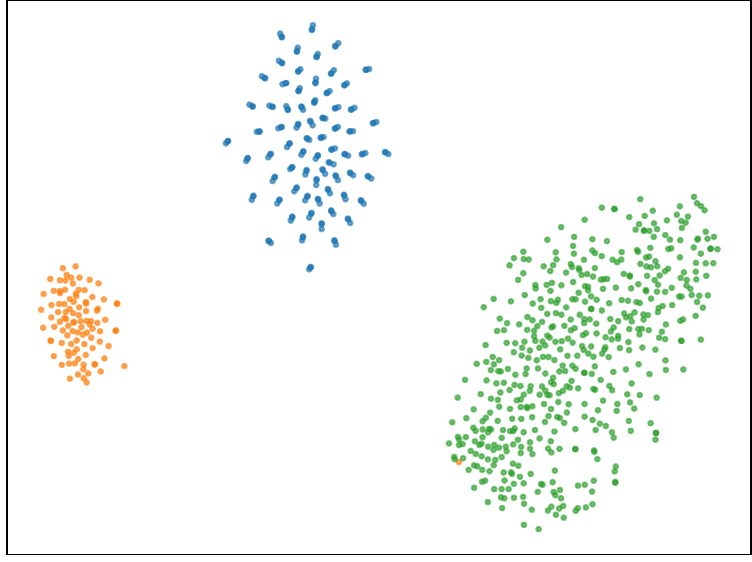}
         &\includegraphics[height=0.1\linewidth,width = 0.145\textwidth]{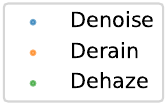}\\   
         AirNet~\cite{airnet}& PromptIR~\cite{promptir}&Ours&\\
        \end{tabular}
    \caption{The t-SNE results of intermediate features produced by the three-task all-in-one models. Our model is better at learning discriminative degradation contexts.}
    \label{fig:tsne3d}
    % \vspace{-0.4cm}
\end{figure}

In this paper, we propose an adaptive all-in-one image restoration framework based on frequency mining and modulation. Specifically, the frequency mining module extracts different frequency signals from the input features, guided by an adaptive spectra decomposition of the degraded input image. The extracted features are then refined using a bidirectional module, which facilitates the interactions between different frequency components by exchanging complementary information. Finally, these modulated features are used to transform the original input features via an efficient transposed cross-attention mechanism. With the proposed key design choices, our method can learn discriminative degradation context more effectively than other competing approaches, as shown in Fig.~\ref{fig:tsne3d}. Overall, the following are the main contributions of our work.

\vspace{-1em}
\begin{itemize}
    \item We propose an adaptive all-in-one image restoration framework that leverages both spatial and frequency domain information to effectively decouple degradations from the desired clean image content.
    \item We introduce the Adaptive Frequency Learning Block (AFLB), which is a plugin block specifically designed for easy integration into existing image restoration architectures. The AFLB performs two sequential tasks: firstly, through its Frequency Mining Module (FMiM), it generates low- and high-frequency feature maps via guidance obtained from the spectra decomposition of the original degraded image; secondly, the Frequency Modulation Module (FMoM) within the AFLB calibrates these features by enabling the exchange of information across different frequency bands to effectively handle diverse types of image degradations.
    \item Extensive experiments demonstrate that our \xnet algorithm sets new state-of-the-art performance on several all-in-one image restoration tasks, including image denoising, dehazing, deraining, motion deblurring, and low-light image enhancement.
\end{itemize}

% firstly, through its Frequency Mining Module (FMiM), it dynamically partitions incoming degraded features into low- and high-frequency feature maps
%and achieve adaptive image-specific restoration in terms of frequency according to its input degradation type.

% In summary, the main contributions of this paper are as follows:
% \begin{itemize}
%     \item We propose an adaptive all-in-one image restoration framework that considers by effectively considering the unique frequency-domain requirements of different degradation types.
%     \item We develop a frequency mining module that extracts different frequency features from the input embeddings with the guidance of the adaptively decoupled spectra of the degraded image.
%     \item We present a frequency modulation module that achieves interactions between frequency features based on the characteristics of different frequencies.
%     \item Extensive experiments demonstrate that the proposed model, termed \xnet, achieves state-of-the-art performance on all-in-one image restoration tasks, including image denoising, dehazing, motion deblurring, deraining, and low-light image enhancement.
% \end{itemize}

\section{Related Work}
\textbf{Single-Task Image Restoration.} Image restoration is a fundamental task in computer vision that aims to reconstruct a clean image from its degraded counterpart. Since it is a highly ill-posed problem, many conventional methods have been proposed that utilize hand-crafted features and assumptions to reduce the solution space~\cite{berman2016non,he2010single}. Such solutions, though perform well on some datasets, may not generalize well to complicated real-world images~\cite{zhang2022deep}. Recently, with the rapid advancements in deep learning, a great number of convolutional neural network (CNN) based methods have been proposed and attained superior performance over traditional methods on various image restoration tasks, such as image denoising~\cite{dncnn,ffdnet}, dehazing~\cite{qin2020ffa,mscnn}, deraining~\cite{mspfn,ren2019progressive}, and motion deblurring~\cite{MIMO,cui2023dual}. To model long-range dependencies, Transformer models have been introduced to low-level tasks and significantly advanced state-of-the-art performance~\cite{dehamer,dehazeformer,Tsai2022Stripformer}. Despite the obtained promising performance, these task-specific methods lack generalization beyond certain degradation types and levels. For general image restoration, several network design-based approaches are proposed, which perform favorably on different restoration tasks ~\cite{wang2022uformer,liang2021swinir,grl,restormer}. Although these networks demonstrate robust performance on various restoration tasks, they require training separate copies on different datasets and tasks. Furthermore, applying a separate model for each possible degradation is resource-intensive, and often impractical for deployment, especially on edge devices.

\vspace{0.5em}
\noindent\textbf{All-in-One Image Restoration.} All-in-one image restoration methods can address numerous degradations within a single model~\cite{promptir,yang2023language,jiang2023autodir,chen2023always}. Early unified models ~\cite{ipt,li2020all} employ distinct encoder and decoder heads to attend to different restoration tasks. However, these non-blind methods need prior knowledge about the degradation involved in the corrupted image in order to channelize it to the relevant restoration head. To achieve blind all-in-one image restoration, AirNet~\cite{airnet} learns the degradation representation from the corrupted images using the contrastive learning strategy, and the learned representation is then used to restore the clean image. The subsequent method, IDR~\cite{idr}, models different degradations depending on the underlying physics principles and achieves all-in-one image restoration in two stages. Recently, several prompt-learning-based schemes have been proposed~\cite{promptir,ma2023prores,conde2024high,ai2023multimodal}. For instance, PromptIR~\cite{promptir} presets a series of tunable prompts to encode discriminative information about degradation types, which involve a large number of parameters. Different from the above-mentioned methods, which operate only in the spatial domain, this paper presents an all-in-one image restoration algorithm that exploits information both in spatial and frequency domains.

\begin{figure}[t]
    \centering
    \includegraphics[width=1\linewidth]{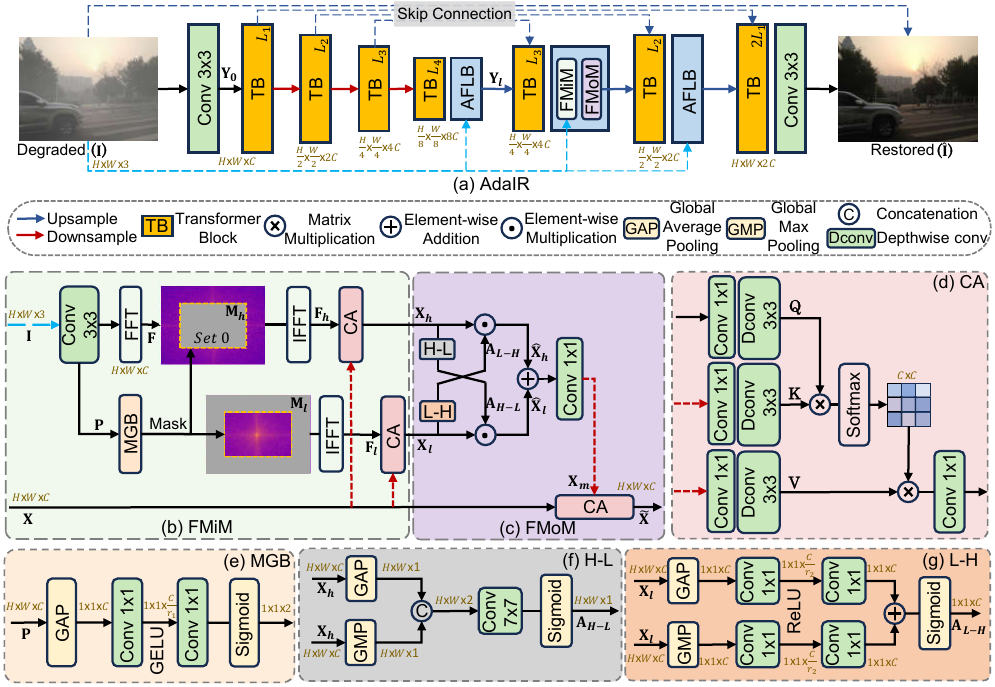}
    \caption{(a) The overall pipeline of the proposed \xnet framework. It is a Transformer-based encoder-decoder architecture, employing novel Adaptive Frequency Learning Blocks (AFLB). Each AFLB contains (b) Frequency Mining Module (FMiM) that extracts different frequency components from input features guided by the adaptively decoupled spectra of the degraded input image, and (c) Frequency Modulation Module (FMoM) that exchanges the complementary information between different frequency features. (d) 
    Cross Attention (CA). (e) Mask Generation Block (MGB) that yields a learnable frequency boundary for spectra decomposition. (f) H-L unit delivers high-frequency attention maps to enrich Low-frequency features. (g) L-H unit enhances high-frequency features by complementing it with low-frequency features. FFT and IFFT denote the Fast Fourier Transform and its inverse operator, respectively.}
    \label{fig:framework}
    % \vspace{-0.3cm}
\end{figure}

\section{Method}
\noindent \textbf{Overall Pipeline.}
Figure~\ref{fig:framework} presents the pipeline of \xnet. The overall goal of our \xnet framework is to learn a unified model $\textbf{M}$ that can recover a clean image $\hat{\textbf{I}}$ from a given degraded image $\textbf{I}$, without any prior information of degradation type $\textbf{D}$ present in the input image $\textbf{I}$.
Formally, given a degraded image $\textbf{I}$~$\in\mathbb{R}^{H\times W\times 3}$, \xnet first extracts shallow features $\mathbf{Y_0}$~$\in$~$\mathbb{R}^{H\times W \times C}$ using a ${3\times 3}$ convolution layer; where ${H\times W}$ denotes the spatial size and $C$ represents the number of channels. Next, these features $\mathbf{Y_0}$ are processed through a 4-level encoder-decoder network. Each level of the encoder employs multiple Transformer blocks (TBs)~\cite{restormer}, where the number of blocks gradually increases from the top level to the bottom level, facilitating a computationally efficient design. The encoder takes high-resolution features $\mathbf{Y_0}$ as input, and progressively transforms them into a lower-resolution latent representation $\mathbf{Y}_l$~$\in$~$\mathbb{R}^{\frac{H}{8}\times\frac{W}{8}\times 8C}$.
On the decoder side, the latent features $\mathbf{Y}_l$ are processed with interleaved Adaptive Frequency Learning Block (AFLB) and TBs to progressively reconstruct high-resolution clean output.
Particularly, between every two levels of the decoder, we insert the AFLB that adaptively segregates the degradation content from the clean image content in the frequency domain, and subsequently assists in refining features in the spatial domain for effective image restoration. 
%

%
% These  are then fed into four sets of TBs to produce in-depth features. During this process, the resolution of features is reduced progressively while the capacity of the channel dimension is expanded, which is implemented by a combination of a $3\times 3$ convolution layer and a pixel-unshuffle operator. After that, the yielded in-depth features pass through interleaved three AFLBs and four groups of TBs to reconstruct high-resolution sharp features. In this process, the resolution of features is gradually restored to that of the input image while the number of channels is reduced from $8C$ to $2C$, which is accomplished using a $3\times 3$ convolution and a pixel-shuffle operation. In addition, the decoder features are concatenated with the corresponding encoder features using skip connections to convey intact spatial signals, which is followed by a $1\times 1$ convolution to adjust the channels, except the top level. These convolutions are omitted in Figure~\ref{fig:framework} (a) for brevity. Subsequently, a $3\times 3$ convolution produces the learned residual image, to which the degraded input is added to attain the final output sharp prediction of the network. Next, we present two key components of AFLB, \textit{i.e.,} FMiM and FMoM. 

% \subsection{Adaptive Frequency Learning Block (AFLB)}
% In degraded
% As illustrated in Figure~\ref{fig:motivation}, different tasks focus on reconstructing different frequency subbands. This observation motivates us 

Since different types of degradations affect image content at different frequency bands (as shown in Fig.~\ref{fig:motivation}), we specifically design the Adaptive Frequency Learning Block (AFLB) that extracts low- and high-frequency components from the input features and then modulate them to accentuate the corresponding informative subbands for each degradation. 
Next, we describe the two key components of AFLB: (1) \textbf{F}requency \textbf{Mi}ning \textbf{M}odule (FMiM) and \textbf{F}requency \textbf{Mo}dulation \textbf{M}odule (FMoM). 
% \vspace{-0.1cm}
% \vspace{0.5em}
% \noindent\textbf{Frequency Mining Module (FMiM).}
\subsection{Frequency Mining Module (FMiM)}
As shown in Fig.~\ref{fig:framework}(b), given as inputs both the degraded image $\textbf{I}$ and the intermediate features $\textbf{X}\in\mathbb{R}^{H\times W\times C}$, FMiM mines different frequency representations from $\textbf{X}$ with the guidance of adaptively decoupled spectra of $\textbf{I}$. 
%, which involves the degradation knowledge. 
%
% The architectural details of FMiM are presented in the left part of Figure~\ref{fig:framework}(b). Taking as input the corrupted image $\textbf{I}$ and intermediate features $\textbf{X}\in\mathbb{R}^{H\times W\times C}$, FMiM mines different frequency parts from $\textbf{X}$ with the guidance of adaptively decoupled spectra of $\textbf{I}$, which involves the degradation knowledge. 
%
Primarily, FMiM consists of three steps, \ie, domain transformation, mask generation, and feature extraction. 

For the domain transformation, FMiM applies a $3\times 3$ convolution layer on the degraded image $\textbf{I}$ to expand the channel capacity to align with that of the input features $\textbf{X}$. These output features are then transformed into spectral domain representation  $\textbf{F}\in\mathbb{R}^{H\times W\times C}$ via the Fast Fourier Transform (FFT). 
% The domain transformation process can be expressed as:
% \begin{equation}
%     \textbf{F}=\mathcal{F}(\textbf{W}^{3\times 3}_{1}(\textbf{I}))
% \end{equation}
% \begin{equation}
% \begin{aligned}
%     \textbf{F}&=\mathcal{F}(\textbf{P}), \\\textbf{P}&=\textbf{W}^{3\times 3}_{1}(\textbf{I}),
%     \end{aligned}
% \end{equation}
% where $\textbf{W}^{3\times 3}_{1}$ indicates a $3\times 3$ convolution layer, $\mathcal{F}$ represents FFT, and $\textbf{F}\in\mathbb{R}^{H\times W\times C}$ is the generated spectra. 
 
Since we want to adaptively extract different frequency parts from the input features $\textbf{X}$, we design a lightweight Mask Generation Block (MGB) to generate a 2D mask that serves as a frequency boundary to separate the spectra of input image $\textbf{I}$. The cutoff frequency boundary adaptively changes according to the type of degradation present in the image. As illustrated in Fig.~\ref{fig:framework}(e), the projected feature map $\textbf{P}$ is first mapped into a vector using a global average pooling operator and then passes through two $1\times 1$ convolution layers with the GELU activation function in between to produce two factors ranging from 0 to 1, which define the mask size by multiplying with the width and height of the spectra. The mask generation process can be formally expressed as:
\begin{equation}
    [\alpha,\beta]= \delta\left(W^{1\times 1}_{2}\left(\sigma\left(W^{1\times 1}_{1}\left(\textrm{GAP}_\textrm{s}\left(\textbf{P}\right)\right)\right)\right)\right)
\end{equation}
where $\textrm{GAP}_\textrm{s}$ denotes spatial global average pooling, $\sigma$ represents the GELU activation function, and $\delta$ indicates the sigmoid function. The convolution weights ${W}_{1}$ and ${W}_{2}$ have the reduction ratios of $r_{1}$ and $\frac{C}{2r_{1}}$, respectively, progressively downsampling the channel dimensions to 2. Subsequently, the binary mask $\textbf{M}_{l}\in\{0,1\}^{H\times W}$ for extracting low frequency can be obtained by setting $\textbf{M}_{l}[\frac{H}{2}-\alpha\frac{H}{k}:\frac{H}{2}+\alpha\frac{H}{k},\frac{W}{2}-\beta\frac{W}{k}:\frac{W}{2}+\beta\frac{W}{k}]=1$, where $k$ is set to a small value of 128, as the curve junction is relatively small in Fig.~\ref{fig:motivation}. Accordingly, the mask for high frequency $\textbf{M}_{h}$ can be obtained by setting the values within the remaining region as 1.
Subsequently, we can obtain the adaptively decoupled features by applying the learned masks to the spectra via element-wise multiplication and using the inverse Fourier transform. 
% which can be expressed as:
% \begin{equation}
% \begin{aligned}
%         \textbf{F}_{l}&=\mathcal{F}^{-1}(\textbf{M}_{l}\odot\textbf{F}),\\
%         \textbf{F}_{h}&=\mathcal{F}^{-1}(\textbf{M}_{h}\odot\textbf{F}),
% \end{aligned}
% \end{equation}
% where $\textbf{F}_{l}$ and $\textbf{F}_{h}$ are the low-frequency and high-frequency features, respectively, $\mathcal{F}^{-1}$ denotes the inverse Fourier transform, and $\odot$ indicates the element-wise multiplication.

Next, we adapt the multi-dconv head transposed cross attention (Fig.~\ref{fig:framework}(d))~\cite{restormer,chen2021crossvit} to mine the different feature parts from the input features with the guidance of $\textbf{F}_{l}$ and $\textbf{F}_{h}$. Overall, the feature extraction process is defined as:
% \begin{equation} `
\begin{gather}
\setlength{\abovedisplayskip}{3pt}
\setlength{\belowdisplayskip}{3pt}
\textbf{X}_{*}=\textrm{softmax}\left(\textbf{Q}\textbf{K}^\top/\alpha\right)\textbf{V}, \quad \quad \textrm{where,}\\ 
    \textbf{Q}={DW}_{1}\left({W}_{3}^{1\times 1}(\textbf{F}_{*})\right),\textbf{K}={DW}_{2}\left({W}_{4}^{1\times 1}(\textbf{X})\right),\textbf{V}={DW}_{3}\left({W}_{5}^{1\times 1}(\textbf{X})\right), \textrm{where,}\\
    \textbf{F}_{*}=\mathcal{F}^{-1}\left(\textbf{M}_{*}\odot\textbf{F}\right),
\end{gather}
% \end{equation}
where $*\in\{l,h\}$ is an indicator for low/high frequency, ${DW}$ represents a $3\times 3$ depth-wise convolution, $\odot$ is element-wise multiplication, $\mathcal{F}^{-1}$ indicates the inverse fast Fourier transform, 
\textbf{Q}, \textbf{K} and \textbf{V} are \textit{query}, \textit{key} and \textit{value} projections, respectively, which are separately generated with a sequential application of ${1\times 1}$ convolution and $3\times 3$ depth-wise convolution, and $\alpha$ is a learnable scaling factor to control the magnitude of the dot product result of \textbf{Q} and \textbf{K} before using the softmax function. 
%
% Finally, the frequency-mined features can be obtained as:
% \begin{equation}
% \begin{aligned}
%     \textbf{X}_{h}&=\mathcal{CA}_{h}(\textbf{F}_{h},\textbf{X},\textbf{X}),\\
%     \textbf{X}_{l}&=\mathcal{CA}_{l}(\textbf{F}_{l},\textbf{X},\textbf{X}).
% \end{aligned} 
% \end{equation}
% In our case, the guidance features serve as the query tensors while input features as the key and value tensors for extraction.
% \vspace{-0.5cm}
% \vspace{0.5em}
% \noindent\textbf{Frequency Modulation Module (FMoM).}
% \vspace{-0.2cm}
\subsection{Frequency Modulation Module (FMoM)}
% \vspace{-0.1cm}
We devise FMoM to facilitate the cross interaction between the low-frequency mined features and high-frequency mined features, shown in Fig.~\ref{fig:framework}(c). The goal is to cross complement one type of mined features with the other. For instance, high-frequency features contain edges and fine texture details, and thus we use this information to enrich low-frequency mined features via a super-lightweight spatial attention unit (H-L), depicted in Fig.~\ref{fig:framework}(f). Similarly, the global information present in low-frequency features is passed to the high-frequency branch through the channel attention unit (L-H), illustrated in Fig.~\ref{fig:framework}(g).

% As illustrated in the right part of Figure~\ref{fig:framework}, we devise FMoM to modulate the mined different frequency features by facilitating interactions between them according to their characteristics. As the high-frequency features contain enriched spatial edges and detailed information while the low-frequency features include global signals, and thus we deliver the \textbf{L}igh-frequency information to the \textbf{H}ow-frequency branch using a lightweight channel attention unit (L-H) and a spatial attention unit (H-L) for the opposite direction, illustrated in Figure~\ref{fig:framework} (e) and Figure~\ref{fig:framework} (f), respectively.

\vspace{0.5em}
\textbf{H-L Unit:} This unit computes the spatial attention map from high-frequency mined features that are then used to complement features of the low-frequency branch. The H-L unit leverages two different channel-wise pooling techniques in parallel to produce two single-channel spatial feature maps, each of size $H\times W\times 1$. These maps are then concatenated along the channel dimension. The concatenated features are further refined with a $7\times 7$ convolution, followed by a sigmoid operation to generate the final spatial attention map, \added{which is then used to obtain the modulated low-frequency features via element-wise multiplication. Overall, the process of the H-L Unit is given by:}
\begin{gather}
\setlength{\abovedisplayskip}{3pt}
\setlength{\belowdisplayskip}{3pt}
\hat{\textbf{X}}_{l}=\textbf{X}_{l}\odot\textbf{A}_{H-L}, \quad \quad 
\textrm{where,}\\ 
\textbf{A}_{H-L}=\delta\left(W^{7\times 7}_{6}([\textrm{GAP}_{c}(\textbf{X}_{h}),\textrm{GMP}_{c}(\textbf{X}_{h})])\right),
\end{gather}
where $\textbf{W}_{6}$ has a channel reduction ratio of 2. $\delta$ is the sigmoid function. $\textrm{GAP}_{c}$ and $\textrm{GMP}_{c}$ are the channel-wise global average pooling and max pooling, respectively. $[\cdot,\cdot]$ indicates the concatenation operation.
% \begin{equation}
%     \hat{\textbf{X}}_{l}=\textbf{X}_{l}\odot\textbf{A}_{H-L}
% \end{equation}

\vspace{0.5em}
\textbf{L-H Unit:} It is a dual branch module that processes incoming low-frequency mined features, yielding a feature descriptor that is subsequently used to attend to the high-frequency mined features. Specifically, given the mined low-frequency features $\textbf{X}_{l}\in\mathbb{R}^{H\times W\times C}$, the top branch of the L-H unit applies global average pooling along spatial dimension to obtain a feature vector of size $1\times 1\times C$, followed by two convolutional layers with the ReLU activation function in between. The bottom branch of the L-H unit employs the same structure, with the only difference of Max pooling at the head. The results of the two branches are added together, on which the sigmoid function is applied to produce the final attention descriptor $\textbf{A}_{L-H}\in\mathbb{R}^{1\times 1\times C}$, \added{which is used to modulate the mined high-frequency features $\textbf{X}_{h}$. The process of the L-H Unit is expressed by:}
\begin{gather}
\hat{\textbf{X}}_{h}=\textbf{X}_{h}\odot\textbf{A}_{L-H}, \quad \quad 
\textrm{where,}\\ 
    \textbf{A}_{L-H}=\delta\left(W_{8}^{1\times 1}\left(\gamma\left(W_{7}^{1\times 1}(\textrm{GAP}_{s}(\textbf{X}_{l})))\right)+W_{10}^{1\times 1}\left(\gamma(W_{9}^{1\times 1}(\textrm{GMP}_{s}(\textbf{X}_{l}))\right)\right)\right),
\end{gather}
where $\delta$ is the sigmoid function, $\hat{\textbf{X}}_{h}$ is the modulated high-frequency features, $\textrm{GAP}_{s}$ and $\textrm{GMP}_{s}$ represent the global average pooling and max pooling along the spatial dimensions, respectively. $\gamma$ indicates the ReLU activation function. $\textbf{W}_{7}$ and $\textbf{W}_{9}$ have a reduction ratio of $r_{2}$ for the channel adjustment, while $\textbf{W}_{8}$ and $\textbf{W}_{10}$ have an increasing ratio of $r_{2}$. The parameters are shared among $\textbf{W}_{7}$ and $\textbf{W}_{9}$, $\textbf{W}_{8}$ and $\textbf{W}_{10}$ for computational efficiency.

Subsequently, the modulated high-frequency features $\hat{\textbf{X}}_{h}$ and low-frequency features $\hat{\textbf{X}}_{l}$ are aggregated and processed via a $1\times 1$ convolution to obtain $\textbf{X}_{m}$, which is merged into the original input features \textbf{X} using the cross-attention unit, where the \textit{query} \textbf{Q} tensor is produced from \textbf{X} while $\textbf{X}_{m}$ yields the \textit{key} \textbf{K} and \textit{value} \textbf{V} tensors.
%
% Subsequently, the combined modulated features are processed via a $1\times 1$ convolution layer and merged into the original input features using an above-mentioned cross-attention unit, which can be expressed as:
% \begin{gather}
%     \widetilde{\textbf{X}}=\mathcal{CA}(\textbf{X},\textbf{X}_{lh},\textbf{X}_{lh})\\
%     \textbf{X}_{lh}=\textbf{W}_{9}^{1\times 1}(\hat{\textbf{X}}_{l}+\hat{\textbf{X}}_{h})
% \end{gather}
%
By using FMiM and FMoM, the high-frequency and low-frequency contents of the input features are separately and adaptively modulated according to the degradation type present in the corrupted input image, leading to adaptive all-in-one image restoration.
% \vspace{-0.2cm}
\section{Experiments}

To validate the efficacy of the proposed \xnet, we conduct experiments by strictly following previous state-of-the-art works~\cite{promptir, airnet} under two different settings: \textbf{(1) All-in-One}, and \textbf{(2) Single-task}.
In the All-in-One setting, a unified model is trained to perform image restoration across multiple degradation types. Whereas, within the Single-task setting, separate models are trained for each specific restoration task.
We provide additional ablation experiments, visual examples, and more details on the architecture in the supplementary material.
In tables, the best and second-best image fidelity scores (PSNR and SSIM~\cite{ssim}) are highlighted in \textcolor{red}{red} and \textcolor{blue}{blue}, respectively.

% To verify the efficacy of the proposed \xnet, we conduct experiments by following previous state-of-the-art works~\cite{promptir, idr} under two settings: the three-task and five-task settings. The first one strictly follows~\cite{promptir}, where we train an all-in-one model for three representative image restoration tasks, \textit{i.e.,} dehazing, denoising, and deraining, on compound data and train individual models for each task. In the five-task setting, we train and test the all-in-one model on a mixed dataset obtained from five tasks, including image denoising, dehazing, deblurring, deraining, low-light image enhancement, and further directly evaluate the trained model on single datasets to inspect the generalization ability.

% In this section, we first describe the implementation details and introduce the used datasets and evaluation protocols. Next, we compare our experimental results with state-of-the-art schemes under different settings. Finally, the ablation studies are performed to verify the efficacy of our proposed components. We provide more details of the used datasets, more experimental results and visual comparisons in the supplementary material.

\vspace{0.5em}
\noindent\textbf{Implementation Details.}
Our \xnet presents an end-to-end trainable solution without the necessity for pretraining any individual component. The architecture of \xnet employs a 4-level encoder-decoder structure, with varying numbers of Transformer blocks (TB) at each level, specifically [4, 6, 6, 8] from level-1 to level-4. We integrate one AFLB block between every two consecutive decoder levels, amounting to a total of three AFLBs in the overall network.

For training, we adopt a batch size of 32 in the all-in-one setting, and a batch size of 8 in the single-task setting. The network optimization is achieved through an L1 loss function, employing the Adam optimizer ($\beta1=0.9$ and $\beta2=0.999$), with a learning rate of $2e^{-4}$, over the course of 150 epochs. During the training process, cropped patches sized at ${128\times 128}$ pixels are provided as input, with additional augmentation applied via random horizontal and vertical flips.

% Our \xnet network is trained end-to-end, without pre-training any individual component. The numbers of Transformer blocks TB ($L_{1}$ to $L_{4}$ in Figure~\ref{fig:framework}) are set as [4,6,6,8] and the basic number of channels ($C$ in Figure~\ref{fig:framework}) is 48.
% The model is trained using the Adam optimizer~\cite{adam} with the learning rate of $2e^{-4}$. We adopt the $L_{1}$ loss function for optimization and random horizontal and vertical flips for data augmentation. The batch size is set to 32 for the all-in-one setting and 8 for the single-task setting. All models are trained for 150 epochs on the patch size of $128\times 128\times 3$. The best results in the tables are \underline{underlined}. 

\vspace{0.5em}
\noindent\textbf{Datasets.}\label{sec:datasets}
In preparing datasets for training and testing, we closely follow prior works~\cite{promptir, airnet}. For single-task image dehazing, we use SOTS~\cite{RESIDE}  dataset that comprises 72,135 training images and 500 testing images. For single-task image deraining, we utilize the Rain100L~\cite{rain100L} dataset, which contains 200 clean-rainy image pairs for training and 100 pairs for testing. For single-task image denoising, we combine images of BSD400~\cite{bsd400} and WED~\cite{wed} datasets for model training; the BSD400 encompasses 400 training images, while the WED dataset consists of 4,744 images. Starting from these clean images of BSD400~\cite{bsd400} and WED~\cite{wed}, we generate their corresponding noisy versions by adding Gaussian noise with varying levels ($\sigma\in \{15,25,50\}$). Denoising task evaluation is performed on the BSD68~\cite{bsd68} and Urban100~\cite{urban100} datasets. Finally, under the all-in-one setting, we train a single model on the combined set of the aforementioned training datasets, and directly test it across multiple restoration tasks.

% Our training dataset setup strictly follows the prior arts for the three-task~\cite{airnet,promptir} and five-task~\cite{idr} settings. Specifically, we adopt BSD400~\cite{bsd400} and WED~\cite{wed} for image denoising by adding Gaussian noise with different noise levels $\sigma\in\{15,25,50\}$, Rain100L~\cite{rain100L} for image deraining, and SOTS~\cite{RESIDE} for image dehazing. For the five-task setting, we additionally use GoPro~\cite{gopro} and LOL~\cite{lol} for image motion deblurring and low-light image enhancement, respectively. Except for image denoising using BSD68~\cite{bsd68}, Urban100~\cite{urban100}, and Kodak24~\cite{kodak24} for evaluation, the evaluation on other datasets follows the standard practices in dataset splitting. We measure the performance of models using the peak signal-to-noise ratio (PSNR) and structural similarity index (SSIM)~\cite{ssim}.

\begin{table}[t!]\scriptsize
\caption{Comparisons under the three-degradation all-in-one setting: a unified model is trained on a combined set of images obtained from all degradation types and levels. On Rain100L~\cite{rain100L} for image deraining, AdaIR yields 2.27 dB gain over PromptIR~\cite{promptir}.}

\label{tab:3D}
\centering
\setlength\tabcolsep{3pt}
\resizebox{\textwidth}{!}{
\begin{tabular}{@{}l|c|c|ccc|c@{}}
\toprule
 & Dehazing & Deraining & \multicolumn{3}{c|}{Denoising on BSD68~\cite{bsd68} } &  \\
Method & on SOTS~\cite{RESIDE} & on Rain100L~\cite{rain100L}  & $\sigma=15$ & $\sigma=25$ & $\sigma=50$ & Average \\ \midrule
BRDNet~\cite{brdnet} & 23.23/0.895 & 27.42/0.895 & 32.26/0.898 & 29.76/0.836 & 26.34/0.693 & 27.80/0.843 \\
LPNet~\cite{lpnet} & 20.84/0.828 & 24.88/0.784 & 26.47/0.778 & 24.77/0.748 & 21.26/0.552 & 23.64/0.738 \\
FDGAN~\cite{fdgan} & 24.71/0.929 & 29.89/0.933 & 30.25/0.910 & 28.81/0.868 & 26.43/0.776 & 28.02/0.883 \\
MPRNet~~\cite{Zamir_2021_CVPR_mprnet} & 25.28/0.955 & 33.57/0.954 & 33.54/0.927 & 30.89/0.880 & 27.56/0.779 & 30.17/0.899 \\ 
DL~\cite{dl} & 26.92/0.931 & 32.62/0.931 & 33.05/0.914 & 30.41/0.861 & 26.90/0.740 & 29.98/0.876 \\
AirNet~\cite{airnet} & 27.94/0.962 & 34.90/0.968 & 33.92/0.933 & 31.26/0.888 & 28.00/0.797 & 31.20/0.910 \\
PromptIR~\cite{promptir} & \textcolor{blue}{30.58}/\textcolor{blue}{0.974} & \textcolor{blue}{36.37}/\textcolor{blue}{0.972} & \textcolor{blue}{33.98}/\textcolor{blue}{0.933} & \textcolor{blue}{31.31}/\textcolor{blue}{0.888} & \textcolor{blue}{28.06}/\textcolor{blue}{0.799} & \textcolor{blue}{32.06}/\textcolor{blue}{0.913} \\ \midrule
\textbf{\xnet (Ours)} &\textcolor{red}{31.06}/\textcolor{red}{0.980}  &\textcolor{red}{38.64}/\textcolor{red}{0.983} &\textcolor{red}{34.12}/\textcolor{red}{0.935} & \textcolor{red}{31.45}/\textcolor{red}{0.892} &\textcolor{red}{28.19}/\textcolor{red}{0.802} & \textcolor{red}{32.69}/\textcolor{red}{0.918} \\ \bottomrule
\end{tabular}}
\end{table}

\begin{figure*}[t!]
\tabcolsep 0.8pt
\centering
		\begin{tabular}{cccccc}

        \includegraphics[height=0.13\linewidth,width = 0.16\textwidth]{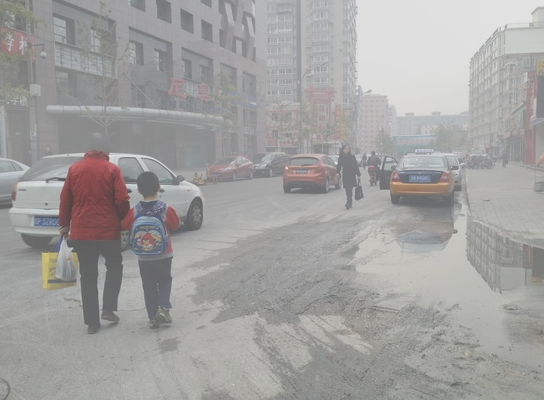}
        &\includegraphics[height=0.13\linewidth,width = 0.16\textwidth]{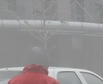}
        &\includegraphics[height=0.13\linewidth,width = 0.16\textwidth]{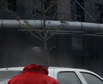}
        &\includegraphics[height=0.13\linewidth,width = 0.16\textwidth]{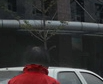} 
        &\includegraphics[height=0.13\linewidth,width = 0.16\textwidth]{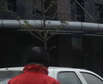} 
        &\includegraphics[height=0.13\linewidth,width = 0.16\textwidth]{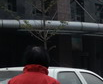} \\
        Degraded &7.84 dB&23.09 dB& 25.30 dB &30.80 dB& PSNR\\

        \includegraphics[height=0.13\linewidth,width = 0.16\textwidth]{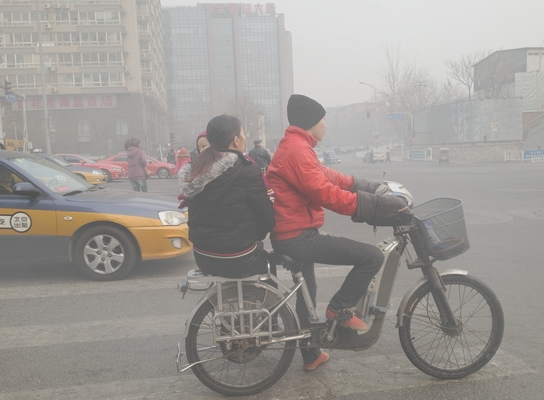}
        &\includegraphics[height=0.13\linewidth,width = 0.16\textwidth]{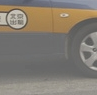}
        &\includegraphics[height=0.13\linewidth,width = 0.16\textwidth]{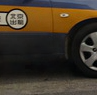}
        &\includegraphics[height=0.13\linewidth,width = 0.16\textwidth]{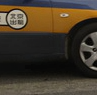} 
        &\includegraphics[height=0.13\linewidth,width = 0.16\textwidth]{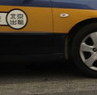} 
        &\includegraphics[height=0.13\linewidth,width = 0.16\textwidth]{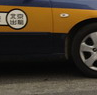} \\
        Degraded &10.82 dB&27.49 dB& 28.75 dB &31.68 dB& PSNR\\

        \includegraphics[height=0.13\linewidth,width = 0.16\textwidth]{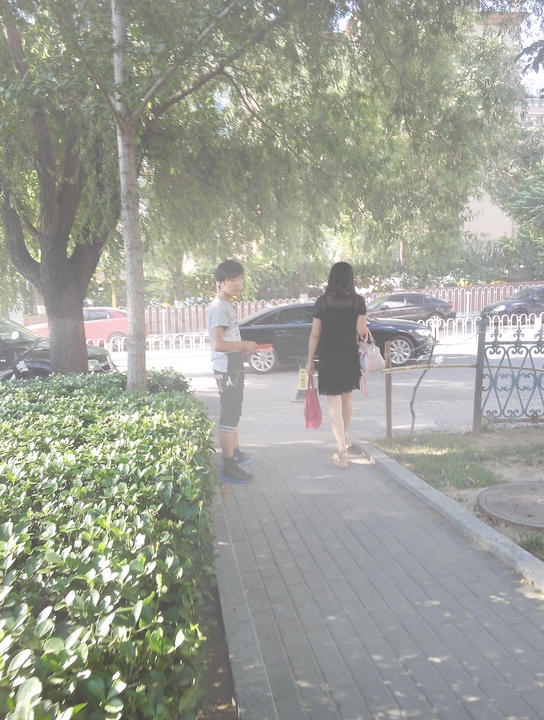}
        &\includegraphics[height=0.13\linewidth,width = 0.16\textwidth]{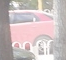}
        &\includegraphics[height=0.13\linewidth,width = 0.16\textwidth]{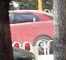}
        &\includegraphics[height=0.13\linewidth,width = 0.16\textwidth]{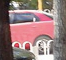} 
        &\includegraphics[height=0.13\linewidth,width = 0.16\textwidth]{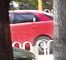} 
        &\includegraphics[height=0.13\linewidth,width = 0.16\textwidth]{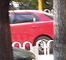} \\
        Degraded &9.57 dB&19.34 dB& 19.88 dB & 24.61 dB& PSNR\\   
        Image & Input& AirNet& PromptIR   &  \textbf{\xnet} & Reference\\
\end{tabular}

\caption{Image dehazing comparisons on SOTS~\cite{RESIDE} between all-in-one methods. Compared to other algorithms, our method is more effective in haze removal.}
\label{fig:3d-hazes}
\end{figure*}

\begin{figure*}[t!]
\tabcolsep 0.8pt
\centering
		\begin{tabular}{cccccc}

        \includegraphics[height=0.13\linewidth,width = 0.16\textwidth]{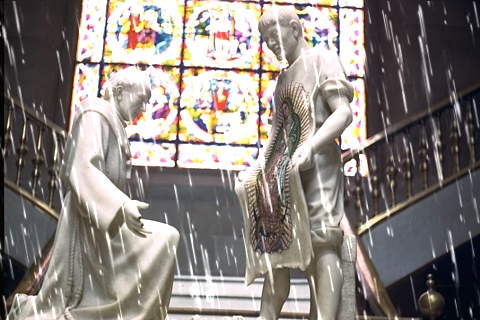}
        &\includegraphics[height=0.13\linewidth,width = 0.16\textwidth]{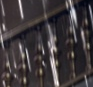}
        &\includegraphics[height=0.13\linewidth,width = 0.16\textwidth]{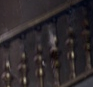}
        &\includegraphics[height=0.13\linewidth,width = 0.16\textwidth]{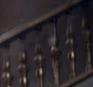} 
        &\includegraphics[height=0.13\linewidth,width = 0.16\textwidth]{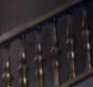} 
        &\includegraphics[height=0.13\linewidth,width = 0.16\textwidth]{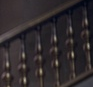} \\
        Degraded &16.92 dB&28.40 dB& 28.81 dB & 31.38 dB& PSNR\\

        \includegraphics[height=0.13\linewidth,width = 0.16\textwidth]{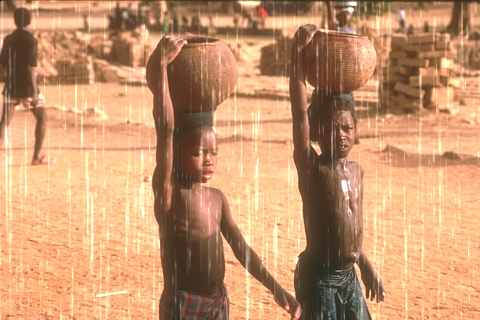}
        &\includegraphics[height=0.13\linewidth,width = 0.16\textwidth]{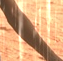}
        &\includegraphics[height=0.13\linewidth,width = 0.16\textwidth]{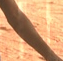}
        &\includegraphics[height=0.13\linewidth,width = 0.16\textwidth]{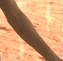} 
        &\includegraphics[height=0.13\linewidth,width = 0.16\textwidth]{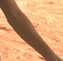} 
        &\includegraphics[height=0.13\linewidth,width = 0.16\textwidth]{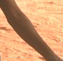} \\
        Degraded &23.16 dB&31.12 dB& 33.64 dB & 38.66 dB& PSNR\\

        \includegraphics[height=0.13\linewidth,width = 0.16\textwidth]{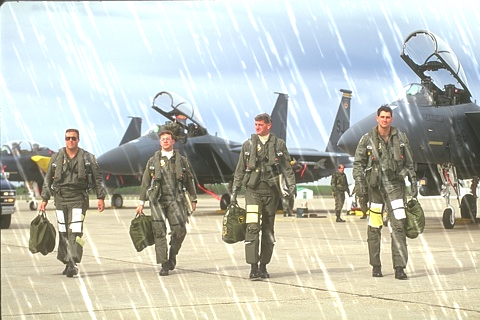}
        &\includegraphics[height=0.13\linewidth,width = 0.16\textwidth]{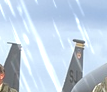}
        &\includegraphics[height=0.13\linewidth,width = 0.16\textwidth]{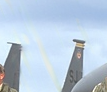}
        &\includegraphics[height=0.13\linewidth,width = 0.16\textwidth]{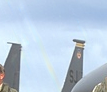} 
        &\includegraphics[height=0.13\linewidth,width = 0.16\textwidth]{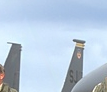} 
        &\includegraphics[height=0.13\linewidth,width = 0.16\textwidth]{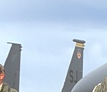} \\
        Degraded &27.49 dB&35.53 dB& 39.25 dB & 44.90 dB& PSNR\\
        Image & Input& AirNet& PromptIR   &  \textbf{\xnet} & Reference\\
\end{tabular}

\caption{Image deraining results on Rain100L~\cite{rain100L} between all-in-one methods. AdaIR yields high-fidelity rain-free images with structural fidelity and without streak artifacts.}
\label{fig:3d-rain}
\end{figure*}

% \vspace{-0.2cm}
\subsection{All-in-One Results: Three Distinct Degradations}

\begin{figure*}[t!]
\tabcolsep 0.8pt
\centering
		\begin{tabular}{cccccc}

        \includegraphics[height=0.13\linewidth,width = 0.16\textwidth]{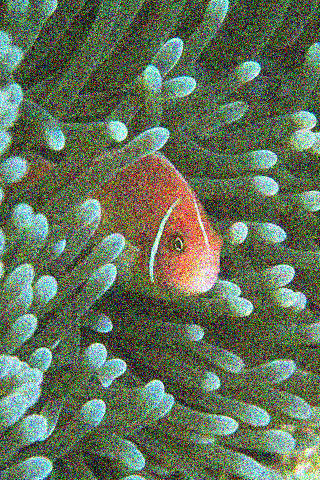}
        &\includegraphics[height=0.13\linewidth,width = 0.16\textwidth]{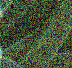}
        &\includegraphics[height=0.13\linewidth,width = 0.16\textwidth]{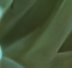}
        &\includegraphics[height=0.13\linewidth,width = 0.16\textwidth]{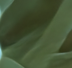} 
        &\includegraphics[height=0.13\linewidth,width = 0.16\textwidth]{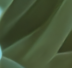} 
        &\includegraphics[height=0.13\linewidth,width = 0.16\textwidth]{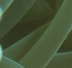} \\
        Degraded &14.95 dB&33.11 dB& 32.91 dB & 34.02 dB& PSNR\\

        \includegraphics[height=0.13\linewidth,width = 0.16\textwidth]{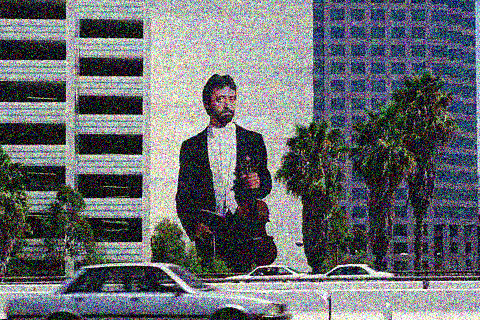}
        &\includegraphics[height=0.13\linewidth,width = 0.16\textwidth]{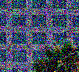}
        &\includegraphics[height=0.13\linewidth,width = 0.16\textwidth]{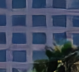}
        &\includegraphics[height=0.13\linewidth,width = 0.16\textwidth]{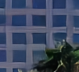} 
        &\includegraphics[height=0.13\linewidth,width = 0.16\textwidth]{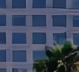} 
        &\includegraphics[height=0.13\linewidth,width = 0.16\textwidth]{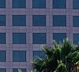} \\
        Degraded &14.89 dB&27.19 dB& 27.23 dB & 27.68 dB& PSNR\\

        \includegraphics[height=0.13\linewidth,width = 0.16\textwidth]{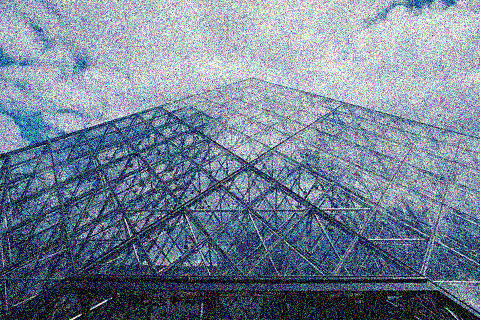}
        &\includegraphics[height=0.13\linewidth,width = 0.16\textwidth]{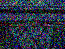}
        &\includegraphics[height=0.13\linewidth,width = 0.16\textwidth]{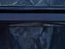}
        &\includegraphics[height=0.13\linewidth,width = 0.16\textwidth]{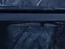} 
        &\includegraphics[height=0.13\linewidth,width = 0.16\textwidth]{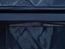} 
        &\includegraphics[height=0.13\linewidth,width = 0.16\textwidth]{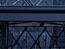} \\
        Degraded &15.63 dB&26.73 dB& 26.18 dB & 27.12 dB& PSNR\\

        Image & Input& AirNet& PromptIR   &  \textbf{\xnet} & Reference\\
\end{tabular}
\vspace{-0.5em}
\caption{Image denoising comparisons on BSD68~\cite{bsd68} between all-in-one methods. The image reproduction quality of our AdaIR is more visually faithful to the ground truth.}
\label{fig:3d-noise}
\end{figure*}

We evaluate the performance of our \emph{all-in-one} \xnet model on three different restoration tasks, including image dehazing, deraining, and denoising. We compare \xnet against various general image restoration methods (BRDNet~\cite{brdnet}, LPNet~\cite{lpnet}, FDGAN~\cite{fdgan}, and MPRNet~\cite{Zamir_2021_CVPR_mprnet}), as well as specialized all-in-one approaches (DL~\cite{dl}, AirNet~\cite{airnet}, and PromptIR~\cite{promptir}). 
%The data configuration follows that of~\cite{airnet,promptir}. 
Table~\ref{tab:3D} shows that the proposed \xnet provides consistent performance gains over the other competing approaches. 
When averaged across various restoration tasks and settings, our \xnet obtains $0.63$ dB PSNR gain over the recent best method PromptIR~\cite{promptir}, and $1.49$ dB improvement over the second best algorithm AirNet~\cite{airnet}. 
Specifically, compared to PromptIR~\cite{promptir}, \xnet yields a substantial boost of $2.27$ dB on the deraining task, and $0.48$ dB on the dehazing task. 
We provide visual examples in Fig.~\ref{fig:3d-hazes} for dehazing, Fig.~\ref{fig:3d-rain} for deraining, and Fig.~\ref{fig:3d-noise} for denoising. These examples show that our \xnet is effective in removing degradations, and generates images that are visually closer to the ground truth than those of the other approaches~\cite{promptir, airnet}. Particularly, in the restored images, our method preserves better structural fidelity and fine textures. %\added{in the first example of Fig.~\ref{fig:3d-rain}}, and fine textures \added{in the first two images of Fig.~\ref{fig:3d-noise}}. 
% the proposed method significantly outperforms the recent algorithm PromptIR~\cite{promptir} with performance improvements of 0.43 dB PSNR and 0.005 SSIM when averaged across all degradation types and levels. Specifically on individual tasks, our \xnet obtains performance gains of 0.48 dB and 2.27 dB PSNR over the PromptIR~\cite{promptir} method on the image dehazing and deraining tasks, respectively. The visual comparisons for all-in-one methods on image dehazing, deraining, and denoising are illustrated in Figure~\ref{fig:3d-hazes}, Figure~\ref{fig:3d-rain}, and Figure~\ref{fig:3d-noise}, respectively. Compared to the strong competitors, \textit{i.e.,} AirNet~\cite{airnet} and PromptIR~\cite{promptir}, our model is more effective in removing different kinds of degradations. For example, the derained result of PromptIR remains some streak artifacts, while our prediction is much closer to the rain-free ground truth.
% \vspace{-0.2cm}
\subsection{Single Degradation One-by-One Results}
% \vspace{-0.1cm}
Consistent with previous works~\cite{airnet,promptir}, we further evaluate \xnet under the \emph{single-task} experimental protocol. To this end, we train separate copies of \xnet model for each distinct restoration task. Table~\ref{tab:sots} reports dehazing results; compared to the previous all-in-one approaches PromptIR~\cite{promptir} and AirNet~\cite{airnet}, our method obtains PSNR gains of $0.49$ dB and $8.62$ dB, respectively. Similarly, on the deraining task, our \xnet advances the state-of-the-art~\cite{promptir} by $1.86$ dB as shown in Table~\ref{tab:rain100l}. A similar performance trend can be observed in image quality scores provided in Table~\ref{tab:bsd68} for denoising.

\begin{table*}[t!]\scriptsize
    \centering
    \caption{Dehazing results in the single-task setting on the SOTS-Outdoor~\cite{RESIDE} dataset. Compared to PromptIR~\cite{promptir}, our method generates a 0.49 dB PSNR improvement.}
    \label{tab:sots}
    \setlength\tabcolsep{2.5pt}
    \begin{tabular}{@{}l|ccccccccc}
    \toprule
         & DehazeNet & MSCNN & AODNet & EPDN &FDGAN & AirNet & Restormer &PromptIR & \textbf{\xnet} \\
    Method & \cite{cai2016dehazenet} &\cite{mscnn} &\cite{li2017aod}&\cite{epdn}&\cite{fdgan}&\cite{airnet}&\cite{restormer}&\cite{promptir}&\textbf{(Ours)}\\  \midrule
    PSNR  &22.46 & 22.06 & 20.29 & 22.57 & 23.15 & 23.18 & 30.87 & \textcolor{blue}{31.31} & \textcolor{red}{{31.80}}  \\
    SSIM& 0.851 & 0.908 & 0.877 & 0.863 & 0.921 & 0.900 & 0.969 & \textcolor{blue}{0.973} &\textcolor{red}{{0.981}} \\ \bottomrule
    \end{tabular}
\end{table*}

\begin{table}[t]\scriptsize
\setlength\tabcolsep{4.1pt}
\centering
\caption{Deraining results in the single-task setting on the Rain100L~\cite{rain100L} dataset. Our AdaIR obtains a significant performance boost of 1.86 dB PSNR over PromptIR~\cite{promptir}.}
\label{tab:rain100l}
\begin{tabular}{@{}l|ccccccccc@{}}
\midrule
 & DIDMDN & UMR & SIRR & MSPFN & LPNet & AirNet & Restormer & PromptIR & \textbf{\xnet} \\
Method & \cite{didmdn} & \cite{umr} &\cite{sirr} &\cite{mspfn} &\cite{lpnet} & \cite{airnet} & \cite{restormer} & \cite{promptir}  &\textbf{(Ours)} \\ \midrule
PSNR & 23.79 & 32.39 & 32.37 & 33.50 & 33.61 & 34.90 & 36.74 & \textcolor{blue}{37.04} & \textcolor{red}{{38.90}} \\
SSIM & 0.773 & 0.921 & 0.926 & 0.948 & 0.958 & 0.977 & 0.978 & \textcolor{blue}{0.979} & \textcolor{red}{{0.985}} \\ \midrule
\end{tabular}
\vspace{-0.2cm}
\end{table}

\begin{table}[t!]\scriptsize
\setlength\tabcolsep{0.6pt}
\centering
\caption{Denoising results in the single-task setting on  Urban100~\cite{urban100} and BSD68~\cite{bsd68}. On Urban100~\cite{urban100} for the noise level 50, AdaIR yields a 0.31 dB gain over PromptIR~\cite{promptir}.}
\label{tab:bsd68}
% \rowcolors{3}{blue!6}{white}
\begin{tabular}{@{}l|c|c|c|c|c|c|c@{}}
\toprule
 & \multicolumn{3}{c|}{Urban100} &\multicolumn{3}{c|}{BSD68}  &  \\ 
Method & $\sigma=15$ & $\sigma=25$ & $\sigma=50$ & $\sigma=15$ & $\sigma=25$ & $\sigma=50$ & Average\\ \midrule
CBM3D~\cite{cbm3d} & 33.93/0.941 & 31.36/0.909 & 27.93/0.840 & 33.50/0.922 & 30.69/0.868 & 27.36/0.763  & 30.80/0.874\\
DnCNN~\cite{dncnn}  & 32.98/0.931 & 30.81/0.902 & 27.59/0.833& 33.89/0.930 & 31.23/0.883 & 27.92/0.789&30.74/0.878 \\
IRCNN~\cite{zhang2017learning}& 27.59/0.833 & 31.20/0.909 & 27.70/0.840& 33.87/0.929 & 31.18/0.882 & 27.88/0.790 &29.90/0.864 \\
FFDNet~\cite{ffdnet} & 33.83/0.942 & 31.40/0.912 & 28.05/0.848& 33.87/0.929 & 31.21/0.882 & 27.96/0.789 &31.05/0.884 \\
BRDNet~\cite{brdnet}& 34.42/0.946 & 31.99/0.919 & 28.56/0.858  & 34.10/0.929 & 31.43/0.885 & 28.16/0.794 &31.44/0.889\\
AirNet~\cite{airnet} & 34.40/0.949 & 32.10/0.924 & 28.88/0.871& 34.14/0.936 & 31.48/0.893 & 28.23/0.806 &31.54/0.897\\
PromptIR~\cite{promptir}& \textcolor{blue}{34.77}/\textcolor{blue}{0.952} & \textcolor{blue}{32.49}/\textcolor{blue}{0.929} & \textcolor{blue}{29.39}/\textcolor{blue}{0.881} & \textcolor{blue}{34.34}/\textcolor{blue}{0.938} & \textcolor{blue}{31.71}/\textcolor{blue}{0.897} & \textcolor{blue}{28.49}/\textcolor{blue}{0.813}  &\textcolor{blue}{31.87}/\textcolor{blue}{0.902}\\ \midrule
% FrePrompter & 34.33/\underline{0.938} & 31.67/\underline{0.897} & 28.42/0.812 & 34.86/0.953 & 32.53/0.930 & 29.36/0.881 \\ \midrule
\textbf{\xnet (Ours)}& \textcolor{red}{34.96}/\textcolor{red}{0.953} &\textcolor{red}{32.74}/\textcolor{red}{0.931}   &  \textcolor{red}{29.70}/\textcolor{red}{0.885} &\textcolor{red}{34.36}/\textcolor{red}{0.938} &\textcolor{red}{31.72}/\textcolor{red}{0.897} & \textcolor{red}{28.49}/\textcolor{red}{0.813}    & \textcolor{red}{32.00}/\textcolor{red}{0.903} \\ \bottomrule
\end{tabular}
\vspace{-0.1cm}
\end{table}

\begin{table}[t!]\scriptsize
\setlength\tabcolsep{3pt}
\centering
\caption{Comparisons for five-degradation all-in-one restoration. Denoising results are reported for the noise level $\sigma=25$. The top super-row methods denote the general image restoration approaches, and the rest are specialized all-in-one approaches. On SOTS~\cite{rain100L} for dehazing, AdaIR attains a remarkable gain of 5.29 dB over IDR~\cite{idr}.}
\label{tab:5D}
\resizebox{\textwidth}{!}{
\begin{tabular}{@{}l|c|c|c|c|c|c@{}}
\toprule
 & Dehazing & Deraining & Denoising & Deblurring& Low-Light  &  \\ 
 Method & on SOTS & on Rain100L &on BSD68 & on GoPro & on LOL &Average\\ \midrule
NAFNet~\cite{chen2022simple}&25.23/0.939 &35.56/0.967 &31.02/0.883 &26.53/0.808 &20.49/0.809& 27.76/0.881 \\
HINet~\cite{hinet}&24.74/0.937& 35.67/0.969 & 31.00/0.881& 26.12/0.788& 19.47/0.800 &27.40/0.875 \\
MPRNet~\cite{Zamir_2021_CVPR_mprnet}&24.27/0.937& \textcolor{red}{38.16}/\textcolor{blue}{0.981}  &31.35/\textcolor{blue}{0.889} &26.87/0.823& 20.84/0.824 &28.27/0.890\\
DGUNet~\cite{dgunet}& 24.78/0.940& 36.62/0.971  & 31.10/0.883&  27.25/0.837 & \textcolor{blue}{21.87}/0.823 & 28.32/0.891\\
MIRNetV2~\cite{mirnet}& 24.03/0.927 &33.89/0.954 &30.97/0.881 &26.30/0.799 &21.52/0.815 &27.34/0.875\\
SwinIR~\cite{liang2021swinir}&21.50/0.891& 30.78/0.923  &30.59/0.868& 24.52/0.773 &17.81/0.723& 25.04/0.835 \\
Restormer~\cite{restormer}& 24.09/0.927 &34.81/0.962 &\textcolor{blue}{31.49}/0.884 &27.22/0.829 &20.41/0.806 &27.60/0.881 \\  \midrule
DL~\cite{dl}&20.54/0.826 &21.96/0.762 & 23.09/0.745 &19.86/0.672 &19.83/0.712& 21.05/0.743\\
Transweather~\cite{transweather}&21.32/0.885& 29.43/0.905 & 29.00/0.841& 25.12/0.757& 21.21/0.792& 25.22/0.836\\
TAPE~\cite{tape}&22.16/0.861 &29.67/0.904  &30.18/0.855 &24.47/0.763 &18.97/0.621 &25.09/0.801\\
AirNet~\cite{airnet} &21.04/0.884 &32.98/0.951 &30.91/0.882 &24.35/0.781 &18.18/0.735 &25.49/0.846\\
IDR~\cite{idr}&\textcolor{blue}{25.24}/\textcolor{blue}{0.943}&35.63/0.965 & \textcolor{red}{31.60}/0.887 &\textcolor{blue}{27.87}/\textcolor{blue}{0.846} &21.34/\textcolor{blue}{0.826} &\textcolor{blue}{28.34}/\textcolor{blue}{0.893} \\  \midrule
\textbf{\xnet (Ours)} & \textcolor{red}{30.53}/\textcolor{red}{0.978}  & \textcolor{blue}{38.02}/\textcolor{red}{0.981}& 31.35/\textcolor{red}{0.889} & \textcolor{red}{28.12}/\textcolor{red}{0.858}&\textcolor{red}{23.00}/\textcolor{red}{0.845} &  \textcolor{red}{30.20}/\textcolor{red}{0.910} \\ \bottomrule
\end{tabular}}
\end{table}

\begin{table}[t]\scriptsize
\setlength\tabcolsep{2.1pt}
\centering
\caption{Image denoising results of directly applying the pre-trained model under the five-degradation setting to the Urban100~\cite{urban100}, Kodak24~\cite{kodak24} and BSD68~\cite{bsd68} datasets. The results are PSNR scores. On Urban100~\cite{urban100} for the noise level $\sigma=25$, AdaIR produces a significant performance gain of 0.39 dB PSNR over IDR~\cite{idr}.}
\label{tab:kodak24}
\begin{tabular}{@{}l|ccc|ccc|ccc|c@{}}
\toprule
& \multicolumn{3}{c|}{Urban100}& \multicolumn{3}{c|}{Kodak24}  & \multicolumn{3}{c|}{BSD68} \\ 
 Method & $\sigma=15$ &$\sigma=25$&$\sigma=50$ &  $\sigma=15$ &$\sigma=25$&$\sigma=50$ & $\sigma=15$ &$\sigma=25$&$\sigma=50$& Average\\ \midrule
DL~\cite{dl}&21.10& 21.28& 20.42& 22.63 &22.66& 21.95& 23.16 &23.09& 22.09  &22.04\\
Transweather~\cite{transweather}&29.64& 27.97& 26.08& 31.67& 29.64 &26.74&31.16 &29.00 &26.08 &28.66\\
TAPE~\cite{tape} &32.19 &29.65& 25.87 &33.24 &30.70& 27.19&32.86 &30.18 &26.63 &29.83\\
AirNet~\cite{airnet} &33.16 &30.83 &27.45& 34.14 &31.74 &28.59&33.49& 30.91& 27.66 &30.89\\
IDR~\cite{idr} &\textcolor{blue}{33.82}& \textcolor{blue}{31.29} &\textcolor{blue}{28.07}& \textcolor{blue}{34.78}& \textcolor{red}{32.42}& \textcolor{blue}{29.13}&\textcolor{red}{34.11}& \textcolor{red}{31.60} &\textcolor{red}{28.14}&\textcolor{blue}{31.48}\\  \midrule
\textbf{\xnet (Ours)} & \textcolor{red}{34.10} & \textcolor{red}{31.68} & \textcolor{red}{28.29} &\textcolor{red}{34.89} & \textcolor{blue}{32.38} & \textcolor{red}{29.21} &  \textcolor{blue}{34.01} & \textcolor{blue}{31.35} & \textcolor{blue}{28.06} &\textcolor{red}{31.55}\\ \bottomrule
\end{tabular}
\end{table}

\subsection{Additional All-in-One Results: Five Distinct Degradations}
Following the recent work of IDR~\cite{idr}, we further verify the effectiveness of \xnet by performing experiments on five restoration tasks: dehazing, deraining, denoising, deblurring, and low-light image enhancement. For this, we train an all-in-one \xnet model on combined datasets gathered for five different tasks. These include datasets from the aforementioned three-task setting as well as additional datasets: GoPro~\cite{gopro} for motion deblurring, and LOL-v1~\cite{lol} for low-light image enhancement.
%

% , we provide experimental results for the five-task setting. We compare our results with representative general image restoration algorithms, including NAFNet~\cite{chen2022simple}, HINet~\cite{hinet}, MPRNet~\cite{Zamir_2021_CVPR_mprnet}, DGUNet~\cite{dgunet}, MIRNetV2~\cite{mirnet}, SwinIR~\cite{liang2021swinir}, and Restormer~\cite{restormer}, and all-in-one algorithms, including DL~\cite{dl}, Transweather~\cite{transweather}, TAPE~\cite{tape}, AirNet~\cite{airnet}, and IDR~\cite{idr}.
% The data setup is consistent with that of IDR~\cite{idr}. 
Table~\ref{tab:5D} shows that AdaIR achieves a $1.86$ dB gain compared to the recent best method IDR~\cite{idr}, when averaged across five restoration tasks. Particularly, the performance improvement is over $5$ dB on dehazing. Table~\ref{tab:kodak24} reports denoising results on three different datasets with various noise levels. It can be seen that our method performs favorably well compared to the other competing approaches. 

% % by an average score of $1.86$ dB PSNR. Particularly on the image dehazing task, our model achieves remarkable performance gains of 5.29 dB PSNR and 0.035 SSIM, demonstrating the effectiveness of our model in removing different degradations.

% In addition, we directly apply the trained all-in-one model to two other image denoising datasets and compare the results with all-in-one algorithms. As shown in Table~\ref{tab:kodak24}, our method performs favorably against state-of-the-art approaches across three datasets.

\begin{table}[t]\scriptsize
\centering
\tabcolsep 3.5pt
\caption{Ablation studies for the proposed components. \textit{Fixed} uses a fixed square mask with sides of 10. FLOPs are measured on the patch size of $256\times 256\times 3$.}
\label{tab:ablations}
% \rowcolors{3}{gray!20}{white}
\begin{tabular}{lccccc|cccc}
\toprule
 & \multicolumn{1}{l}{} & \multicolumn{2}{c}{FMiM} & \multicolumn{2}{c}{FMoM} & \multicolumn{1}{l}{} & \multicolumn{1}{l}{} & \multicolumn{2}{c}{Overhead} \\
Net & \multicolumn{1}{|c}{Baseline} & \multicolumn{1}{|c}{Fixed} & MGB & \multicolumn{1}{|c}{L-H} & H-L & \multicolumn{1}{c}{PSNR} & \multicolumn{1}{c|}{SSIM} & Params. & FLOPs \\ \midrule
(a) & \Checkmark  &  &  &  &  & 28.21 &0.966  & 26.13M & 141.24G \\
(b) & \Checkmark & \Checkmark &  &  &  & 29.79 & 0.969 &27.73M  &145.09G  \\
(c) & \Checkmark & \Checkmark &  & \Checkmark &  & 30.37 &0.975  & 28.74M &147.44G  \\
(d) & \Checkmark &\Checkmark  &  & \Checkmark &\Checkmark  &30.52  & 0.976 & 28.74M & 147.44G \\
(e) & \Checkmark &  & \Checkmark & \Checkmark & \Checkmark & 31.24 & 0.978 & 28.77M &  147.45G\\ \bottomrule
\end{tabular}
\end{table}

\subsection{Ablation Studies}
In this section, we conduct ablation studies to test the impact of various individual components to the overall performance of \xnet. 
%We carry out ablation studies to demonstrate the effectiveness of the proposed modules and investigate different alternatives. 
All ablation experiments are performed on the image dehazing task by training models for 20 epochs.

\vspace{0.5em}
\noindent\textbf{Impact of individual architecture modules.} 
Table~\ref{tab:ablations} summarizes the performance benefits of individual architectural contributions. Table~\ref{tab:ablations}(b) demonstrates that the proposed frequency mining mechanism (FMiM) brings gains of $1.58$ dB PSNR over the baseline model, using only a fixed mask to decompose the spectra of input images. 
 Furthermore, the L-H unit boosts the performance to $30.37$ dB PSNR; see Table~\ref{tab:ablations}(c). It can be seen in Table~\ref{tab:ablations}(d) that we use both L-H and H-L units, and the performance reaches $30.52$ dB PSNR. Finally, Table~\ref{tab:ablations}(e) shows that the overall \xnet brings $3.03$ dB improvement over the baseline, while incurring a small computational overhead of 2.64M parameters and 6.21 GFlops. These results corroborate the effectiveness of our design.

%This ablation investigates the influence of the proposed archite. The results are presented in Table~\ref{tab:ablations}. The baseline model is obtained by removing our modules from \xnet, achieving 28.21 dB PSNR on the SOTS~\cite{RESIDE} dataset. The proposed frequency mining mechanism achieves a performance gain of 1.58 dB PSNR over the baseline model, using only a fixed mask to decompose the spectra of input images (Table~\ref{tab:ablations}b). Moreover, L-H boosts the performance to 30.37 dB PSNR (Table~\ref{tab:ablations}c) while the combination of L-H and H-L achieves a further improvement to 30.52 dB PSNR (Table~\ref{tab:ablations}d). The complete model, additionally using a learned mask to decouple the spectra via MGB, produces the best performance, which only introduces extra 2.64M parameters and 6.21 GFLOPs. These results suggest the effectiveness of our design.

\vspace{0.5em}
\noindent\textbf{Strategies for spectral decomposition.} 
We carry out this ablation to test different strategies to segregate low- and high-frequency representations from the degraded input images. We compare the proposed mask-guided adaptive frequency decomposition approach with the Average pooling and Gaussian filtering strategies. Results are provided in Table~\ref{tab:spectra}. Following ~\cite{cui2023focal}, we use average pooling to obtain the low-frequency features which are then subtracted from the input features to obtain the high-frequency features. This strategy provides PSNR of $30.59$ (see column 1 in Table~\ref{tab:spectra}), which is  $0.65$ dB lower than our method. Similarly, when we switch to the Gaussian filter of size ${5\times 5}$, the model achieves only $30.22$ dB PSNR (second column). In contrast, our method of applying a flexible mask for Fourier spectra decomposition performs the best, yielding $31.24$ dB.

\begin{table*}[t!]\scriptsize
\centering
\begin{minipage}[c]{0.5\textwidth}\scriptsize
\caption{Spectra decomposition methods.}
\label{tab:spectra}
\centering
\setlength\tabcolsep{2.6pt}
\begin{tabular}{l|ccc}
\toprule
Methods &Average Pooling & Gaussian Filter & Ours \\ \midrule
  PSNR& 30.59 &   30.22 & 31.24  \\
 SSIM &0.976 & 0.976 & 0.978\\ \bottomrule
% \rowcolor{gray!20}
\end{tabular}
\end{minipage}~~
\begin{minipage}[c]{0.48\textwidth}\scriptsize
\caption{Degradation sources.}
\label{tab:space}
\centering
\setlength\tabcolsep{8pt}
\begin{tabular}{l|ccc|c}
\toprule
Method &  Embedding & Ours \\ \midrule
PSNR& 29.29 &  30.52 \\
% \rowcolor{gray!20}
 SSIM& 0.969& 0.976 \\ \bottomrule
\end{tabular}
\end{minipage}
% \begin{minipage}[c]{0.27\textwidth}\scriptsize
% \caption{Design options for FMoM.}
% \label{tab:direction}
% \centering
% \setlength\tabcolsep{4pt}
% \begin{tabular}{l|ccc|c}
% \toprule
% Method & PSNR & SSIM\\ \midrule
% Inverse & 30.38 & 0.975 \\
% % \rowcolor{gray!20}
% Ours & 30.52 & 0.976 \\ \bottomrule
% \end{tabular}
% \end{minipage}
\end{table*}

\begin{table*}[t!]\scriptsize
\centering
\begin{minipage}[c]{0.49\textwidth}\scriptsize
\caption{Results on the unseen desnowing task with the CSD~\cite{csd} dataset.}
\label{tab:csd}
\centering
\setlength\tabcolsep{3pt}
\begin{tabular}{l|ccc}
\toprule
Method & AirNet~\cite{airnet} & PromptIR~\cite{promptir} &Ours\\ \midrule
 PSNR&19.32 & 20.47  &  20.54 \\
 SSIM & 0.733 & 0.7638 &0.7643\\\bottomrule
\end{tabular}
\end{minipage}~
\begin{minipage}[c]{0.49\textwidth}\scriptsize
\caption{Results on mixed degradations, Rain100L with the Gaussian noise $\sigma=50$.}
\label{tab:mixed}
\centering
\setlength\tabcolsep{3pt}
\begin{tabular}{l|ccc}
\toprule
Method & AirNet~\cite{airnet} & PromptIR~\cite{promptir} & Ours \\ \midrule
PSNR & 27.25 & 27.34 &27.51 \\
SSIM & 0.790 & 0.791 &0.799 \\ \bottomrule
\end{tabular}
\end{minipage}
\end{table*}

\vspace{0.5em}
\noindent\textbf{Frequency representation mining at image-level vs. feature-level.}
Each AFLB block in \xnet decoder receives the original degraded image as input, on which FMiM applies the procedure of spectra decomposition. To verify the efficacy of this design, we switch to using the input embedding features $\mathbf{X}$ (rather than degraded image) for frequency representation. This ablation result in Table~\ref{tab:space} shows a performance drop from $30.52$ dB to $29.29$ dB, indicating that the raw input image offers better discriminative information about the degradation for effective spectra separation.

% \textbf{Design choices for FMoM.} FMoM adopts H-L to deliver detailed spatial information from the high-frequency branch to the low-frequency branch, and L-H for the opposite direction to convey global information. As shown in Table~\ref{tab:direction}, reversing the directions of these two units leads to a 0.14 dB lower PSNR than ours, revealing the effectiveness of our design.

\vspace{0.5em}
\noindent\textbf{Generalization to out-of-distribution degradations.} To show the generalization ability of our \xnet, we take the all-in-one model trained on the three-task setting, and directly test it under two different scenarios: (1) unseen degradation type, and (2) multi-degraded images. 
Table~\ref{tab:csd} shows that, on the unseen task of image desnowing, \xnet provides more favorable results than other approaches.  
%
% Next, we evaluate the model to a highly-challenging unseen noise level of $\sigma=100$. Table~\ref{tab:sigma35} shows that our method generalizes better than PromptIR~\cite{promptir} by yielding $????$ dB improvement. 
%
We create a mixed degradation dataset by adding Gaussian noise (level $\sigma=50$) to the rainy images of Rain100L~\cite{rain100L}. Table~\ref{tab:mixed} depicts that our method is more robust in the mixed degradation scenes than PromptIR~\cite{promptir} and AirNet~\cite{airnet}.

%by applying the pre-trained three-task all-in-one model to unseen degradation levels and tasks. Specifically, we first apply the model to an unseen noise level $\sigma=35$. Table~\ref{tab:sigma35} shows that our method significantly outperforms the strong competitor PromptIR~\cite{promptir} by 0.15 dB and 0.41 dB PSNR on the BSD68~\cite{bsd68} and Urban100~\cite{urban100} datasets, respectively. We further evaluate the model, trained on image dehazing, deraining and denoising datasets, on the out-of-distribution image denoising task with the CSD~\cite{csd} dataset. Table~\ref{tab:csd} shows 
% the superiority of our generalization ability over PromptIR~\cite{promptir} by outperforming it with 0.05 dB PSNR. Finally, we create a mixed degradation dataset by imposing Gaussian noise with the noise level $\sigma=50$ on the rainy images of Rain100L~\cite{rain100L}. Table~\ref{tab:mixed} shows that our method is more robust in the mixed degradation scenes than PromptIR~\cite{promptir}. These results reveal that our method has a strong ability to adapt to unseen degradations, which is an essential factor for practical applications.

\section{Conclusion}
This paper introduces \xnet, an all-in-one image restoration model capable of adaptively removing different kinds of image degradations. 
%Our design is motivated by the observation that different degradations affect distinct frequency bands
%different degradation types pay different attention to distinct frequency segments. 
Motivated by the observation that different degradations affect distinct frequency bands, we have developed two novel components: a frequency mining module and a frequency modulation module. 
These modules are designed to identify and enhance the relevant frequency components based on the degradation patterns present in the input image. 
%Based on this, we develop a frequency mining module and a frequency modulation module to recover the informative frequency components with the guidance of the input degraded image.
Specifically, the frequency mining module extracts specific frequency elements from the image's intermediate features, guided by an adaptive decomposition of the input's spectral characteristics that reflect the underlying degradation. 
%the frequency mining module extracts different frequency parts from the intermediate features guided by the adaptively decomposed spectra of the input image that involves the degradation information. 
Subsequently, the frequency modulation module further refines these elements by facilitating the exchange of complementary information across different frequency features.
%the mined different frequency features are refined using the frequency modulation module by exchanging complementary information across different frequency features.
Incorporating the proposed modules into a U-shaped Transformer backbone, the proposed network achieves state-of-the-art performance on a range of image restoration tasks.

\bibliographystyle{splncs04}
\bibliography{eccv}

\begin{thebibliography}{10}
\providecommand{\url}[1]{\texttt{#1}}
\providecommand{\urlprefix}{URL }
\providecommand{\doi}[1]{https://doi.org/#1}

\bibitem{ai2023multimodal}
Ai, Y., Huang, H., Zhou, X., Wang, J., He, R.: Multimodal prompt perceiver: Empower adaptiveness, generalizability and fidelity for all-in-one image restoration. arXiv:2312.02918  (2023)

\bibitem{bsd400}
Arbelaez, P., Maire, M., Fowlkes, C., Malik, J.: Contour detection and hierarchical image segmentation. TPAMI  (2010)

\bibitem{ba2016layer}
Ba, J.L., Kiros, J.R., Hinton, G.E.: Layer normalization. arXiv:1607.06450  (2016)

\bibitem{berman2016non}
Berman, D., Avidan, S., et~al.: Non-local image dehazing. In: CVPR (2016)

\bibitem{brown2020language}
Brown, T., Mann, B., Ryder, N., Subbiah, M., Kaplan, J.D., Dhariwal, P., Neelakantan, A., Shyam, P., Sastry, G., Askell, A., et~al.: Language models are few-shot learners. NeurIPS  (2020)

\bibitem{cai2016dehazenet}
Cai, B., Xu, X., Jia, K., Qing, C., Tao, D.: Dehazenet: An end-to-end system for single image haze removal. TIP  (2016)

\bibitem{chen2021crossvit}
Chen, C.F.R., Fan, Q., Panda, R.: Crossvit: Cross-attention multi-scale vision transformer for image classification. In: ICCV (2021)

\bibitem{ipt}
Chen, H., Wang, Y., Guo, T., Xu, C., Deng, Y., Liu, Z., Ma, S., Xu, C., Xu, C., Gao, W.: Pre-trained image processing transformer. In: CVPR (2021)

\bibitem{chen2022simple}
Chen, L., Chu, X., Zhang, X., Sun, J.: Simple baselines for image restoration. In: ECCV (2022)

\bibitem{hinet}
Chen, L., Lu, X., Zhang, J., Chu, X., Chen, C.: Hinet: Half instance normalization network for image restoration. In: CVPR Workshops (2021)

\bibitem{csd}
Chen, W.T., Fang, H.Y., Hsieh, C.L., Tsai, C.C., Chen, I., Ding, J.J., Kuo, S.Y., et~al.: All snow removed: Single image desnowing algorithm using hierarchical dual-tree complex wavelet representation and contradict channel loss. In: ICCV (2021)

\bibitem{chen2023always}
Chen, Y.W., Pei, S.C.: Always clear days: Degradation type and severity aware all-in-one adverse weather removal. arXiv:2310.18293  (2023)

\bibitem{MIMO}
Cho, S.J., Ji, S.W., Hong, J.P., Jung, S.W., Ko, S.J.: Rethinking coarse-to-fine approach in single image deblurring. In: ICCV (2021)

\bibitem{conde2024high}
Conde, M.V., Geigle, G., Timofte, R.: High-quality image restoration following human instructions. arXiv:2401.16468  (2024)

\bibitem{cui2023focal}
Cui, Y., Ren, W., Cao, X., Knoll, A.: Focal network for image restoration. In: ICCV (2023)

\bibitem{cui2023dual}
Cui, Y., Tao, Y., Ren, W., Knoll, A.: Dual-domain attention for image deblurring. In: AAAI (2023)

\bibitem{cbm3d}
Dabov, K., Foi, A., Katkovnik, V., Egiazarian, K.: Color image denoising via sparse 3d collaborative filtering with grouping constraint in luminance-chrominance space. In: ICIP (2007)

\bibitem{dong2020multi}
Dong, H., Pan, J., Xiang, L., Hu, Z., Zhang, X., Wang, F., Yang, M.H.: Multi-scale boosted dehazing network with dense feature fusion. In: CVPR (2020)

\bibitem{dong2011image}
Dong, W., Zhang, L., Shi, G., Wu, X.: Image deblurring and super-resolution by adaptive sparse domain selection and adaptive regularization. TIP  (2011)

\bibitem{fdgan}
Dong, Y., Liu, Y., Zhang, H., Chen, S., Qiao, Y.: Fd-gan: Generative adversarial networks with fusion-discriminator for single image dehazing. In: AAAI (2020)

\bibitem{dl}
Fan, Q., Chen, D., Yuan, L., Hua, G., Yu, N., Chen, B.: A general decoupled learning framework for parameterized image operators. TPAMI  (2019)

\bibitem{lpnet}
Gao, H., Tao, X., Shen, X., Jia, J.: Dynamic scene deblurring with parameter selective sharing and nested skip connections. In: CVPR (2019)

\bibitem{dehamer}
Guo, C.L., Yan, Q., Anwar, S., Cong, R., Ren, W., Li, C.: Image dehazing transformer with transmission-aware 3d position embedding. In: CVPR (2022)

\bibitem{he2010single}
He, K., Sun, J., Tang, X.: Single image haze removal using dark channel prior. TPAMI  (2010)

\bibitem{urban100}
Huang, J.B., Singh, A., Ahuja, N.: Single image super-resolution from transformed self-exemplars. In: CVPR (2015)

\bibitem{mspfn}
Jiang, K., Wang, Z., Yi, P., Chen, C., Huang, B., Luo, Y., Ma, J., Jiang, J.: Multi-scale progressive fusion network for single image deraining. In: CVPR (2020)

\bibitem{jiang2023autodir}
Jiang, Y., Zhang, Z., Xue, T., Gu, J.: Autodir: Automatic all-in-one image restoration with latent diffusion. arXiv:2310.10123  (2023)

\bibitem{kim2010single}
Kim, K.I., Kwon, Y.: Single-image super-resolution using sparse regression and natural image prior. TPAMI  (2010)

\bibitem{kopf2008deep}
Kopf, J., Neubert, B., Chen, B., Cohen, M., Cohen-Or, D., Deussen, O., Uyttendaele, M., Lischinski, D.: Deep photo: Model-based photograph enhancement and viewing. ACM TOG  (2008)

\bibitem{lester2021power}
Lester, B., Al-Rfou, R., Constant, N.: The power of scale for parameter-efficient prompt tuning. In: EMNLP (2021)

\bibitem{li2017aod}
Li, B., Peng, X., Wang, Z., Xu, J., Feng, D.: Aod-net: All-in-one dehazing network. In: ICCV (2017)

\bibitem{RESIDE}
Li, B., Ren, W., Fu, D., Tao, D., Feng, D., Zeng, W., Wang, Z.: Benchmarking single-image dehazing and beyond. TIP  (2018)

\bibitem{airnet}
Li, B., Liu, X., Hu, P., Wu, Z., Lv, J., Peng, X.: All-in-one image restoration for unknown corruption. In: CVPR (2022)

\bibitem{li2020all}
Li, R., Tan, R.T., Cheong, L.F.: All in one bad weather removal using architectural search. In: CVPR (2020)

\bibitem{grl}
Li, Y., Fan, Y., Xiang, X., Demandolx, D., Ranjan, R., Timofte, R., Van~Gool, L.: Efficient and explicit modelling of image hierarchies for image restoration. In: CVPR (2023)

\bibitem{liang2021swinir}
Liang, J., Cao, J., Sun, G., Zhang, K., Van~Gool, L., Timofte, R.: {SwinIR}: Image restoration using swin transformer. In: ICCV Workshops (2021)

\bibitem{liu2020trident}
Liu, J., Wu, H., Xie, Y., Qu, Y., Ma, L.: Trident dehazing network. In: CVPR Workshops (2020)

\bibitem{tape}
Liu, L., Xie, L., Zhang, X., Yuan, S., Chen, X., Zhou, W., Li, H., Tian, Q.: Tape: Task-agnostic prior embedding for image restoration. In: ECCV (2022)

\bibitem{ma2023prores}
Ma, J., Cheng, T., Wang, G., Zhang, Q., Wang, X., Zhang, L.: Prores: Exploring degradation-aware visual prompt for universal image restoration. arXiv:2306.13653  (2023)

\bibitem{wed}
Ma, K., Duanmu, Z., Wu, Q., Wang, Z., Yong, H., Li, H., Zhang, L.: Waterloo exploration database: New challenges for image quality assessment models. TIP  (2016)

\bibitem{bsd68}
Martin, D., Fowlkes, C., Tal, D., Malik, J.: A database of human segmented natural images and its application to evaluating segmentation algorithms and measuring ecological statistics. In: ICCV (2001)

\bibitem{michaeli2013nonparametric}
Michaeli, T., Irani, M.: Nonparametric blind super-resolution. In: ICCV (2013)

\bibitem{dgunet}
Mou, C., Wang, Q., Zhang, J.: Deep generalized unfolding networks for image restoration. In: CVPR (2022)

\bibitem{gopro}
Nah, S., Hyun~Kim, T., Mu~Lee, K.: Deep multi-scale convolutional neural network for dynamic scene deblurring. In: CVPR (2017)

\bibitem{nah2022clean}
Nah, S., Son, S., Lee, J., Lee, K.M.: Clean images are hard to reblur: Exploiting the ill-posed inverse task for dynamic scene deblurring. In: ICLR (2022)

\bibitem{promptir}
Potlapalli, V., Zamir, S.W., Khan, S.H., Shahbaz~Khan, F.: Promptir: Prompting for all-in-one image restoration. NeurIPS  (2023)

\bibitem{qin2020ffa}
Qin, X., Wang, Z., Bai, Y., Xie, X., Jia, H.: Ffa-net: Feature fusion attention network for single image dehazing. In: AAAI (2020)

\bibitem{epdn}
Qu, Y., Chen, Y., Huang, J., Xie, Y.: Enhanced pix2pix dehazing network. In: CVPR (2019)

\bibitem{ren2021adaptivedeamnet}
Ren, C., He, X., Wang, C., Zhao, Z.: Adaptive consistency prior based deep network for image denoising. In: CVPR (2021)

\bibitem{ren2019progressive}
Ren, D., Zuo, W., Hu, Q., Zhu, P., Meng, D.: Progressive image deraining networks: A better and simpler baseline. In: CVPR (2019)

\bibitem{mscnn}
Ren, W., Liu, S., Zhang, H., Pan, J., Cao, X., Yang, M.H.: Single image dehazing via multi-scale convolutional neural networks. In: ECCV (2016)

\bibitem{ren2020single}
Ren, W., Pan, J., Zhang, H., Cao, X., Yang, M.H.: Single image dehazing via multi-scale convolutional neural networks with holistic edges. IJCV  (2020)

\bibitem{kodak24}
Rich, F.: Kodak lossless true color image suite. \url{http://r0k. us/graphics/kodak} (1999)

\bibitem{shrivastava2023repository}
Shrivastava, D., Larochelle, H., Tarlow, D.: Repository-level prompt generation for large language models of code. In: ICML (2023)

\bibitem{dehazeformer}
Song, Y., He, Z., Qian, H., Du, X.: Vision transformers for single image dehazing. TIP  (2023)

\bibitem{brdnet}
Tian, C., Xu, Y., Zuo, W.: Image denoising using deep cnn with batch renormalization. Neural Networks  (2020)

\bibitem{timofte2013anchored}
Timofte, R., De~Smet, V., Van~Gool, L.: Anchored neighborhood regression for fast example-based super-resolution. In: ICCV (2013)

\bibitem{Tsai2022Stripformer}
Tsai, F.J., Peng, Y.T., Lin, Y.Y., Tsai, C.C., Lin, C.W.: Stripformer: Strip transformer for fast image deblurring. In: ECCV (2022)

\bibitem{tsai2022banet}
Tsai, F.J., Peng, Y.T., Tsai, C.C., Lin, Y.Y., Lin, C.W.: {BANet}: A blur-aware attention network for dynamic scene deblurring. TIP  (2022)

\bibitem{tu2022maxim}
Tu, Z., Talebi, H., Zhang, H., Yang, F., Milanfar, P., Bovik, A., Li, Y.: {MAXIM}: Multi-axis {MLP} for image processing. In: CVPR (2022)

\bibitem{transweather}
Valanarasu, J.M.J., Yasarla, R., Patel, V.M.: Transweather: Transformer-based restoration of images degraded by adverse weather conditions. In: CVPR (2022)

\bibitem{wang2022uformer}
Wang, Z., Cun, X., Bao, J., Zhou, W., Liu, J., Li, H.: Uformer: A general u-shaped transformer for image restoration. In: CVPR (2022)

\bibitem{ssim}
Wang, Z., Bovik, A.C., Sheikh, H.R., Simoncelli, E.P.: Image quality assessment: from error visibility to structural similarity. TIP  (2004)

\bibitem{lol}
Wei, C., Wang, W., Yang, W., Liu, J.: Deep retinex decomposition for low-light enhancement. arXiv:1808.04560  (2018)

\bibitem{sirr}
Wei, W., Meng, D., Zhao, Q., Xu, Z., Wu, Y.: Semi-supervised transfer learning for image rain removal. In: CVPR (2019)

\bibitem{woo2018cbam}
Woo, S., Park, J., Lee, J.Y., So~Kweon, I.: Cbam: Convolutional block attention module. In: ECCV (2018)

\bibitem{yang2023language}
Yang, H., Pan, L., Yang, Y., Liang, W.: Language-driven all-in-one adverse weather removal. arXiv:2312.01381  (2023)

\bibitem{rain100L}
Yang, W., Tan, R.T., Feng, J., Guo, Z., Yan, S., Liu, J.: Joint rain detection and removal from a single image with contextualized deep networks. TPAMI  (2019)

\bibitem{umr}
Yasarla, R., Patel, V.M.: Uncertainty guided multi-scale residual learning-using a cycle spinning cnn for single image de-raining. In: CVPR (2019)

\bibitem{restormer}
Zamir, S.W., Arora, A., Khan, S., Hayat, M., Khan, F.S., Yang, M.H.: Restormer: Efficient transformer for high-resolution image restoration. In: CVPR (2022)

\bibitem{zamir2020cycleisp}
Zamir, S.W., Arora, A., Khan, S., Hayat, M., Khan, F.S., Yang, M.H., Shao, L.: {CycleISP}: Real image restoration via improved data synthesis. In: CVPR (2020)

\bibitem{zamir2020mirnet}
Zamir, S.W., Arora, A., Khan, S., Hayat, M., Khan, F.S., Yang, M.H., Shao, L.: Learning enriched features for real image restoration and enhancement. In: ECCV (2020)

\bibitem{Zamir_2021_CVPR_mprnet}
Zamir, S.W., Arora, A., Khan, S., Hayat, M., Khan, F.S., Yang, M.H., Shao, L.: Multi-stage progressive image restoration. In: CVPR (2021)

\bibitem{mirnet}
Zamir, S.W., Arora, A., Khan, S., Hayat, M., Khan, F.S., Yang, M.H., Shao, L.: Learning enriched features for fast image restoration and enhancement. TPAMI  (2022)

\bibitem{didmdn}
Zhang, H., Patel, V.M.: Density-aware single image de-raining using a multi-stream dense network. In: CVPR (2018)

\bibitem{idr}
Zhang, J., Huang, J., Yao, M., Yang, Z., Yu, H., Zhou, M., Zhao, F.: Ingredient-oriented multi-degradation learning for image restoration. In: CVPR (2023)

\bibitem{dncnn}
Zhang, K., Zuo, W., Chen, Y., Meng, D., Zhang, L.: Beyond a gaussian denoiser: Residual learning of deep cnn for image denoising. TIP  (2017)

\bibitem{zhang2017learning}
Zhang, K., Zuo, W., Gu, S., Zhang, L.: Learning deep {CNN} denoiser prior for image restoration. In: CVPR (2017)

\bibitem{ffdnet}
Zhang, K., Zuo, W., Zhang, L.: Ffdnet: Toward a fast and flexible solution for cnn-based image denoising. TIP  (2018)

\bibitem{zhang2020deblurring}
Zhang, K., Luo, W., Zhong, Y., Ma, L., Stenger, B., Liu, W., Li, H.: Deblurring by realistic blurring. In: CVPR (2020)

\bibitem{zhang2022deep}
Zhang, K., Ren, W., Luo, W., Lai, W.S., Stenger, B., Yang, M.H., Li, H.: Deep image deblurring: A survey. IJCV  (2022)

\end{thebibliography}

\newpage
\appendix
This supplementary material provides additional ablation studies (Sec.~\ref{ablation}), computational comparisons (Sec.~\ref{complexity}), architectural details of the transformer block (Sec.~\ref{TB}), and additional visual results (Sec.~\ref{visual}).

\section{Additional Ablation Studies}
\label{ablation}
% This section provides additional ablation results using the same experimental configurations as that in the main manuscript.

\noindent \textbf{AFLBs in encoder and decoder?} We run an experiment to assess the feasibility of employing AFLB modules on either the encoder side, decoder side, or both. Table~\ref{tab:enc-dec} shows that utilizing AFLBs in both the encoder and decoder leads to notable performance degradation compared to AFLBs solely integrated into the decoder. 
% \textcolor{red}{Yuning, please replicate the ablation style in table A.2 of PromptIR supplement here.}

\begin{table}[h]\scriptsize
    \centering
    \tabcolsep 5pt
    \caption{Comparisons of image dehazing under the single-task setting: between the use of AFLBs on either the encoder-side, decoder-side, or both.}
    \label{tab:enc-dec}
    \begin{tabular}{l|cc}
    \toprule
         & \multicolumn{2}{c}{Dehazing on SOTS~\cite{RESIDE}} \\ 
        Method & PSNR & SSIM \\ \midrule
        Encoder+Decoder+AFLB & 29.70&0.973 \\
        % Encoder+AFLB & 30.48&0.974\\ \midrule
        AdaIR (Ours) & 30.52&0.976 \\  \bottomrule
    \end{tabular}
\end{table}

\noindent\textbf{Placement of AFLB in the network.} Next, we conduct an ablation experiment to study where to place AFLBs in our hierarchical network. Table~\ref{tab:position} demonstrates that employing only one AFLB (between level 1 and level 2)  leads to a deterioration in the network's performance (29.58 dB in top row). Conversely, integrating AFLBs between every consecutive level of the decoder yields the best performance.
% \textcolor{red}{Yuning, please replicate the ablation style in table 6 of the main PromptIR paper here.}
%for the deployment position of the proposed frequency learning mechanism based on the model of Table 7(b) in the main text. Table~\ref{tab:position} shows that the model achieves better performance as we increase the number of the proposed modules. We then place our mechanism only in the encoder stage and the model achieves a 0.04 dB lower performance than our choice, suggesting that our module plays a more important role in restoring high-quality features than extracting features. We further employ our modules in both the encoder and decoder stages, leading to only 29.70 dB PSNR. Finally, we choose to use the module only in the decoder stage for better performance.

\begin{table}[h]\scriptsize
    \centering
    \tabcolsep 6pt
    \caption{AFLB position. Results are reported on the SOTS~\cite{RESIDE} dataset.}
    \label{tab:position}
    \begin{tabular}{l|cc}
    \toprule
       Method  & PSNR & SSIM \\ \midrule
       Level 2  & 28.58 & 0.973 \\
       Level 2+3 & 29.83 &0.975 \\ 
       Level 2+3+4 & 30.52 & 0.976\\ \bottomrule
    \end{tabular}
\end{table}

% \begin{table}[h]\scriptsize
%     \caption{Ablation studies for the deployment position of our frequency learning mechanism. Decoder 2 means only using the last two AFLB (w/ fixed manner) in the model.}
%     \label{tab:position}
%     \tabcolsep 5pt
%     \centering
%     \begin{tabular}{c|ccc|c|c}
%     \toprule
%        Variant & Decoder 1 & Decoder 2 & Decoder 3 & Encoder & Encoder/Decoder  \\ \midrule
%        PSNR &28.58 & 29.83 &  30.52 &30.48 &29.70   \\
%        SSIM & 0.973 & 0.975 &  0.976& 0.974 &0.973\\ \bottomrule
%     \end{tabular}
% \end{table}

\begin{figure}[t]
    \centering
    \begin{tabular}{cc}
      \includegraphics[scale=1]{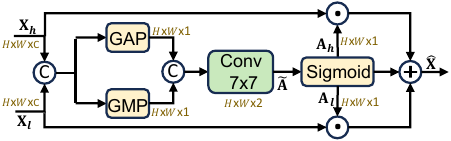}   &   \includegraphics[scale=1]{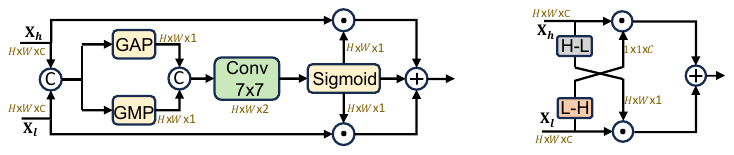}\\
      (a) Spatial attention, 29.67 dB/0.973 & (b) Ours, 30.52 dB/0.976\\
    \end{tabular}
    \caption{Different choices for FMoM. (a) Using widely adopted spatial attention~\cite{woo2018cbam} to modulate different frequency features, where the attention map is generated without discriminating different frequency inputs. (b) Using specially designed attention units to exchange complementary information across different frequency features. GAP and GMP denote the global average pooling and global max pooling, respectively. The experiments are conducted on image dehazing under the single-task setting.} %The results are presented in the form of PSNR/SSIM}
    \label{fig:FMoM-alternative}
\end{figure}

\noindent\textbf{Design choices of FMoM.}
We investigate different choices for the frequency modulation module (FMoM). As shown in Fig.~\ref{fig:FMoM-alternative}(a), we leverage the commonly used spatial attention~\cite{woo2018cbam} to modulate different frequency features without discriminating different inputs. Overall, the process is formally given by:
\begin{gather}
    \hat{\textbf{X}}=\textbf{X}_{h}\odot\textbf{A}_{h}+\textbf{X}_{l}\odot\textbf{A}_{l}, \quad\quad \textrm{where},\\
    \textbf{A}_{h},\textbf{A}_{l}=\textrm{Split}\left(\delta(\widetilde{\textbf{A}})\right),\quad\quad \textrm{where},\\
    \widetilde{\textbf{A}}=W^{7\times 7}\left([\textrm{GAP}([\textbf{X}_{h}, \textbf{X}_{l}]),\textrm{GMP}([\textbf{X}_{h}, \textbf{X}_{l}])]\right)
\end{gather}
where $\odot$ represents element-wise multiplication, Split indicates splitting the features among the channel dimension, $\delta$ is the Sigmoid function, $W^{7\times 7}$ is a $7\times 7$ convolution, and $[\cdot,\cdot]$ is a concatenation operator. GAP and GMP are global average pooling and global max pooling among the channel dimensions, respectively. The experiments are performed on the image dehazing task under the single-task setting.
This variant achieves only 29.67 dB PSNR, which is 0.85 dB lower than our FMoM, shown in Fig.~\ref{fig:FMoM-alternative}(b), indicating the effectiveness of our design.

\vspace{0.5em}
\noindent\textbf{Combinations of different degradations.}
We investigate the influence of various combinations of degradation types on model performance, as presented in Table~\ref{tab:combination}. As expected, including more degradation types make it increasingly difficult for the model to perform restoration. %Specifically, the best results are obtained under the single-task setting for all three degradation types. 
Notable, hazy images in a combined dataset lead to a larger performance drop than rainy or noisy images. One reason could be that the aim of the restoration model in deraining and denoising tasks is to focus more on restoring high-frequency content (noise, rain), whereas, in the dehazing task the goal is to focus on removing low-frequency (hazy) content. %The dehazing task has worse effects on the denoising task than deraining since the deraining and denoising tasks mainly focus on reconstructing high-frequency components while dehazing targets low-frequency signals.

\vspace{0.5em}
\begin{table}[t]\scriptsize
\centering
\setlength\tabcolsep{5.4pt}
\caption{Ablation studies on the combinations of degradations for the three-task setting. Results are presented in the form of PSNR (dB)/SSIM.}
\label{tab:combination}
\rowcolors{3}{gray!10}{white}
\begin{tabular}{ccc|ccc|c|c}
\toprule
\multicolumn{3}{c|}{Degradation} & \multicolumn{3}{c|}{Denoising on BSD68} & Deraining on & Dehazing \\
Noise & Rain & Haze & $\sigma=15$ & $\sigma=25$ & $\sigma=50$ & on Rain100L & on SOTS \\ \midrule
\Checkmark && &34.36/0.938 &31.72/0.897 & 28.49/0.813 & - & - \\
 & \Checkmark &  & - & - & - & 38.90/0.985 & - \\
 &  & \Checkmark & - & - & - & - & 31.80/0.981 \\
\Checkmark & \Checkmark&  & 34.31/0.938 & 31.67/0.896 &28.42/0.811  & 38.22/0.983 &  -\\
\Checkmark &  & \Checkmark & 34.11/0.935 & 31.48/0.892 & 28.19/0.802 & - & 30.89/0.980 \\
 & \Checkmark & \Checkmark & - &-& - &38.44/0.983  & 30.54/0.978\\
\Checkmark & \Checkmark&\Checkmark & 34.12/0.935 & 31.45/0.892 & 28.19/0.802 & 38.64/0.983 & 31.06/0.980 \\ \bottomrule
\end{tabular}
\end{table}

\begin{figure}[t]
    \centering
    \includegraphics[scale=0.7]{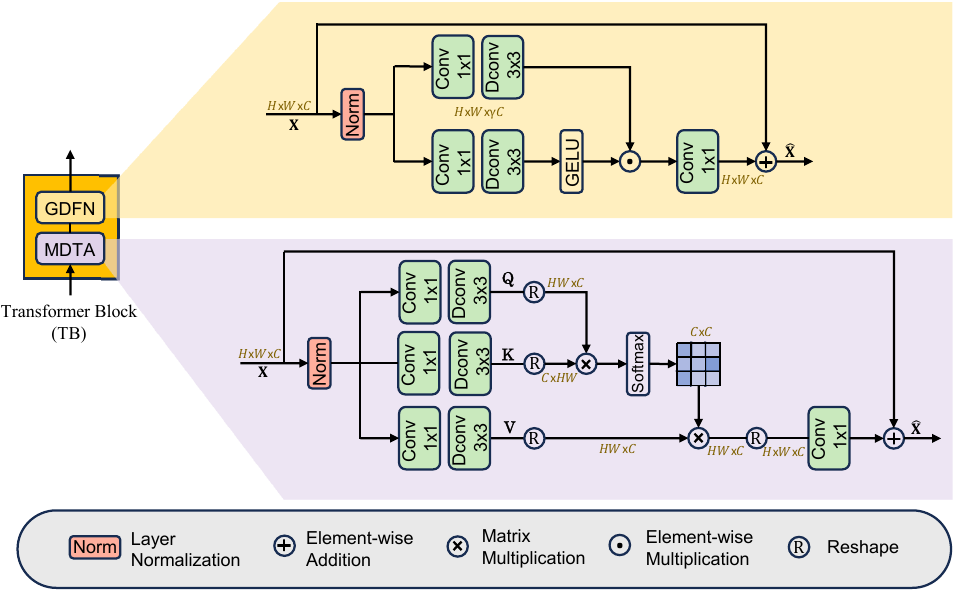}
    \caption{Architectural details of the Transformer Block (TB) used in the AdaIR framework. TB involves two elements: multi-dconv head transposed attention (MDTA) and gated-dconv feed-forward network (GDFN).}
    \label{fig:tb}
\end{figure}

\section{Computational Comparisons}
\label{complexity}
Table~\ref{tab:flops} shows that the proposed AdaIR strikes a better tradeoff between accuracy and complexity than other all-in-one competing methods.

\begin{table}[h]
    \centering
    \caption{Computational comparisons of all-in-one methods under the three-degradation setting. Average PSNR across three tasks is reported here (see Table 1 of the main paper for more detailed results). FLOPs are measured on the patch size of $256\times 256\times3$.}
    \label{tab:flops}
    \tabcolsep 7pt
    \begin{tabular}{l|cccc}
    \toprule
          & Params. & FLOPs  & PSNR  \\
        Method  & (M) & (G)  &\\ \midrule
        AirNet~\cite{airnet} & 8.93 & 311&   31.20 \\
        PromptIR~\cite{promptir} & 35.59 & 158.4& 32.06 \\
        AdaIR & 28.77 & 147.45 & 32.69 \\ \bottomrule
    \end{tabular}
\end{table}

\section{Transformer Block in the AdaIR Framework}
\label{TB}
In the AdaIR framework, we use Transformer Blocks (TB) based on the design proposed in \cite{restormer}.  Fig.~\ref{fig:tb} presents its architectural details. It consists of two successive components, multi-dconv head transposed attention (MDTA) and gated-dconv feed-forward network (GDFN).

MDTA first normalizes the input $\textbf{X}\in\mathbb{R}^{H\times W\times C}$ using a layer normalization operator~\cite{ba2016layer}, and then generates the \textit{query} (\textbf{Q}$\in\mathbb{R}^{H\times W\times C}$), \textit{key} (\textbf{K}$\in\mathbb{R}^{H\times W\times C}$), and \textit{value} (\textbf{V}$\in\mathbb{R}^{H\times W\times C}$) projections using combinations of $1\times 1$ convolution and $3\times 3$ depth-wise convolution layers. The transposed-attention map of size $C\times C$ is yielded by applying the Softmax function to the dot-product results of the reshaped query and key projections. Overall, the process of MDTA is given by:
\begin{gather}
    \hat{\textbf{X}}=W^{1\times 1}_{1}\textrm{Attention}\left(\textbf{Q}^\prime,\textbf{K}^\prime,\textbf{V}^\prime\right)+\textbf{X},\quad \quad\textrm{where,}\\
    \textrm{Attention}\left(\textbf{Q}^\prime,\textbf{K}^\prime,\textbf{V}^\prime\right)=\textbf{V}^\prime\cdot\textrm{Softmax}\left(\textbf{K}^\prime\cdot\textbf{Q}^\prime/\alpha\right),
\end{gather}
where $\hat{\textbf{X}}$ is the output of MDTA. $W_{1}^{1\times 1}$ denotes a $1\times 1$ convolution. $\alpha$ is a learnable factor to control the magnitude of the dot product result of \textbf{K} and \textbf{Q}. $\textbf{Q}^\prime$, $\textbf{K}^\prime$ and $\textbf{V}^\prime$ are obtained by reshaping tensors from the original size $\mathbb{R}^{H\times W\times C}$.

Similarly, GDFN first applies a layer normalization operator to normalize the input $\textbf{X}\in\mathbb{R}^{H\times W\times C}$. The result then passes through two branches, each including a $1\times 1$ convolution with a factor $\gamma$ to expand channels, followed by a $3\times 3$ depth-wise convolution layer. Two branches converge using element-wise multiplication after activating one branch via a GELU function. Overall, the GDFN process is formally expressed as:
\begin{gather}
    \hat{\textbf{X}}=W_{2}^{1\times 1}\textrm{Gating}(\textbf{X})+\textbf{X}, \quad\quad \textrm{where,}\\
    \textrm{Gating}(\textbf{X})=\phi\left(DW^{3\times 3}_{1}\left(W_{3}^{1\times 1}(\textrm{LN}(\textbf{X}))\right)\right)\odot DW^{3\times 3}_{2}\left(W_{4}^{1\times 1}(\textrm{LN}({\textbf{X}}))\right),
\end{gather}
where $\textrm{LN}$ is the layer normalization, $\odot$ denotes element-wise multiplication, $DW^{3\times 3}$ represents a $3\times 3$ depth-wise convolution, and $\phi$ indicates the GELU non-linearity.

\section{Additional Visual Results}
\label{visual}
\begin{figure}[t]
    \centering
    \begin{tabular}{cc}
        \centering
        \includegraphics[scale=0.4]{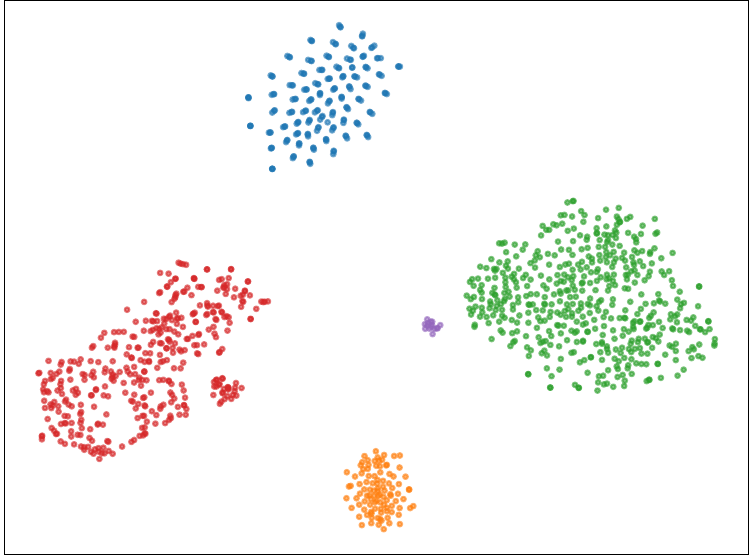}     & 
        \includegraphics[scale=0.55]{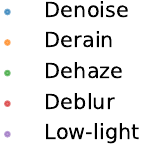}
        \end{tabular}
    \caption{The t-SNE result of our model under the five-degradation setting.}
    \label{fig:tsne-5d}
\end{figure}

In this section, we first provide the t-SNE result of our method under the five-degradation setting in Fig.~\ref{fig:tsne-5d}. It can be seen that our method is capable of discriminating degradation contexts for five different degradation types. It is worth noting that the cluster for low-light image enhancement is closer to the dehazing cluster than others, suggesting the effectiveness of our model, since these two degradation types mainly impact the image content on low-frequency components.

Finally, we provide more qualitative results of the all-in-one setting and single-task setting for three image restoration tasks, including image deraining, dehazing, and denoising.

\begin{figure*}[!htb]
	\begin{center}
		% \tabcolsep 1pt
		% [height=0.24\linewidth,width=0.24\linewidth]
% \resizebox{\linewidth}{!}{
		\begin{tabular}{cccccc}

        \includegraphics[height=0.13\linewidth,width = 0.16\textwidth]{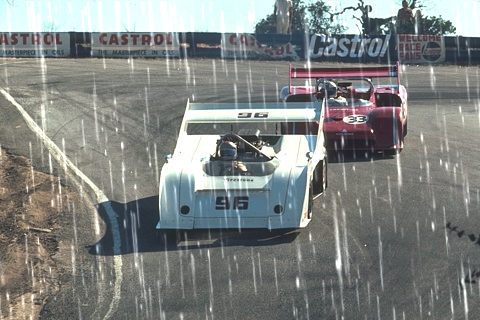}
        &\includegraphics[height=0.13\linewidth,width = 0.16\textwidth]{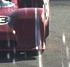}
        &\includegraphics[height=0.13\linewidth,width = 0.16\textwidth]{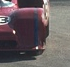}
        &\includegraphics[height=0.13\linewidth,width = 0.16\textwidth]{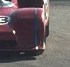} 
        &\includegraphics[height=0.13\linewidth,width = 0.16\textwidth]{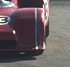} 
        &\includegraphics[height=0.13\linewidth,width = 0.16\textwidth]{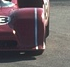} \\
        Degraded &22.95 dB&26.71 dB& 30.03 dB & 34.78 dB& PSNR\\

        \includegraphics[height=0.13\linewidth,width = 0.16\textwidth]{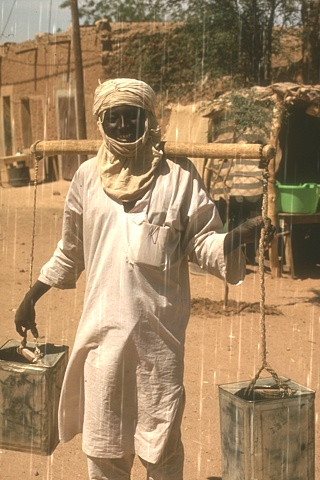}
        &\includegraphics[height=0.13\linewidth,width = 0.16\textwidth]{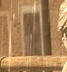}
        &\includegraphics[height=0.13\linewidth,width = 0.16\textwidth]{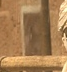}
        &\includegraphics[height=0.13\linewidth,width = 0.16\textwidth]{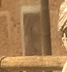} 
        &\includegraphics[height=0.13\linewidth,width = 0.16\textwidth]{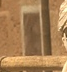} 
        &\includegraphics[height=0.13\linewidth,width = 0.16\textwidth]{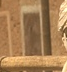} \\
        Degraded &28.61 dB&29.18 dB& 31.11 dB & 35.80 dB& PSNR\\

        \includegraphics[height=0.13\linewidth,width = 0.16\textwidth]{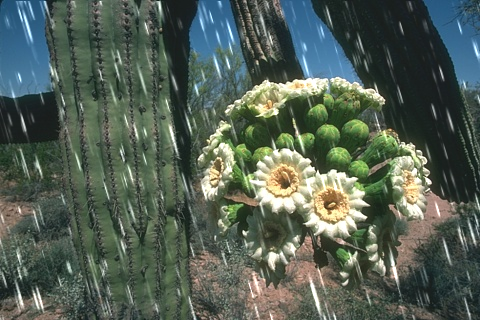}
        &\includegraphics[height=0.13\linewidth,width = 0.16\textwidth]{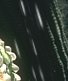}
        &\includegraphics[height=0.13\linewidth,width = 0.16\textwidth]{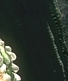}
        &\includegraphics[height=0.13\linewidth,width = 0.16\textwidth]{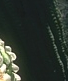} 
        &\includegraphics[height=0.13\linewidth,width = 0.16\textwidth]{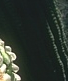} 
        &\includegraphics[height=0.13\linewidth,width = 0.16\textwidth]{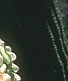} \\
        Degraded &23.08 dB&29.15 dB& 32.00 dB &36.99 dB& PSNR\\

        \includegraphics[height=0.13\linewidth,width = 0.16\textwidth]{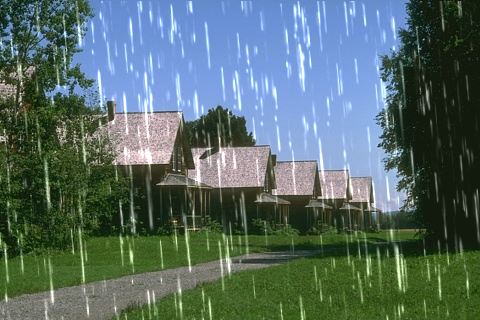}
        &\includegraphics[height=0.13\linewidth,width = 0.16\textwidth]{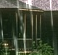}
        &\includegraphics[height=0.13\linewidth,width = 0.16\textwidth]{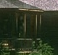}
        &\includegraphics[height=0.13\linewidth,width = 0.16\textwidth]{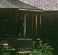} 
        &\includegraphics[height=0.13\linewidth,width = 0.16\textwidth]{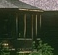} 
        &\includegraphics[height=0.13\linewidth,width = 0.16\textwidth]{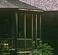} \\
        Degraded &19.97 dB&29.23 dB& 29.74 dB & 32.34 dB& PSNR\\
  
        \includegraphics[height=0.13\linewidth,width = 0.16\textwidth]{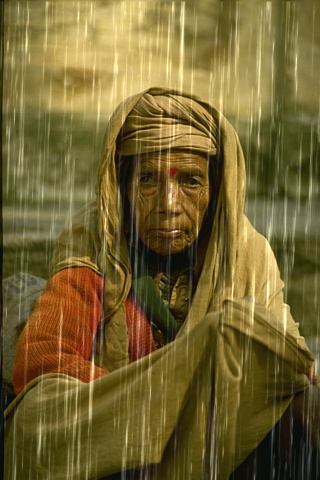}
        &\includegraphics[height=0.13\linewidth,width = 0.16\textwidth]{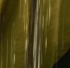}
        &\includegraphics[height=0.13\linewidth,width = 0.16\textwidth]{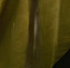}
        &\includegraphics[height=0.13\linewidth,width = 0.16\textwidth]{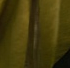} 
        &\includegraphics[height=0.13\linewidth,width = 0.16\textwidth]{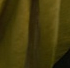} 
        &\includegraphics[height=0.13\linewidth,width = 0.16\textwidth]{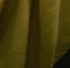} \\
        Degraded &21.09 dB&30.80 dB& 34.17 dB & 36.91 dB& PSNR\\

        \includegraphics[height=0.13\linewidth,width = 0.16\textwidth]{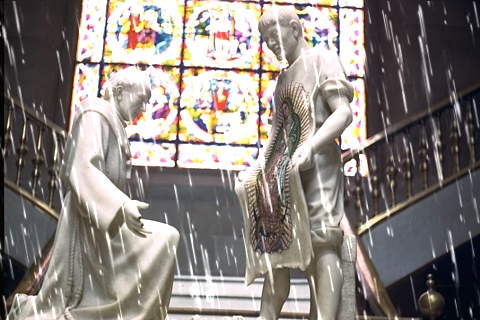}
        &\includegraphics[height=0.13\linewidth,width = 0.16\textwidth]{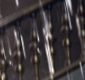}
        &\includegraphics[height=0.13\linewidth,width = 0.16\textwidth]{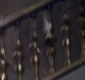}
        &\includegraphics[height=0.13\linewidth,width = 0.16\textwidth]{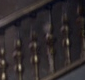} 
        &\includegraphics[height=0.13\linewidth,width = 0.16\textwidth]{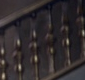} 
        &\includegraphics[height=0.13\linewidth,width = 0.16\textwidth]{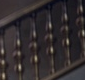} \\
        Degraded &17.49 dB&27.49 dB& 28.22 dB & 30.67 dB& PSNR\\

        \includegraphics[height=0.13\linewidth,width = 0.16\textwidth]{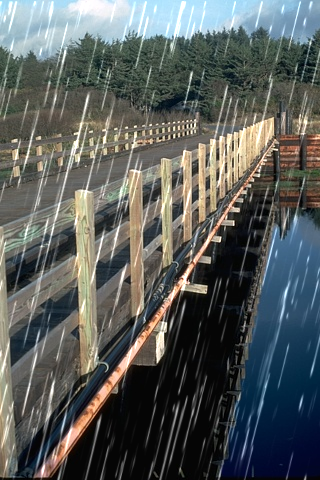}
        &\includegraphics[height=0.13\linewidth,width = 0.16\textwidth]{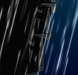}
        &\includegraphics[height=0.13\linewidth,width = 0.16\textwidth]{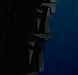}
        &\includegraphics[height=0.13\linewidth,width = 0.16\textwidth]{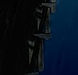} 
        &\includegraphics[height=0.13\linewidth,width = 0.16\textwidth]{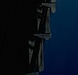} 
        &\includegraphics[height=0.13\linewidth,width = 0.16\textwidth]{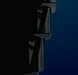} \\
        Degraded &18.00 dB&32.94 dB& 33.61 dB & 37.13 dB& PSNR\\

        \includegraphics[height=0.13\linewidth,width = 0.16\textwidth]{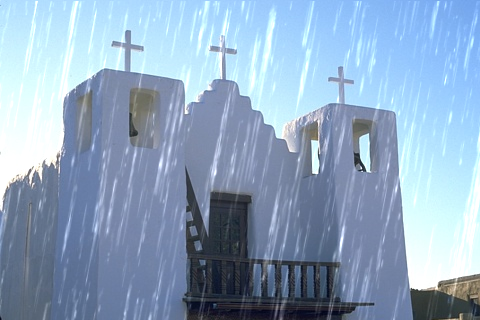}
        &\includegraphics[height=0.13\linewidth,width = 0.16\textwidth]{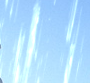}
        &\includegraphics[height=0.13\linewidth,width = 0.16\textwidth]{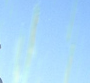}
        &\includegraphics[height=0.13\linewidth,width = 0.16\textwidth]{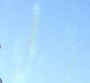} 
        &\includegraphics[height=0.13\linewidth,width = 0.16\textwidth]{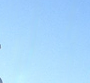} 
        &\includegraphics[height=0.13\linewidth,width = 0.16\textwidth]{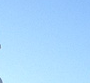} \\
        Degraded &25.47 dB&33.46 dB& 36.80 dB & 44.37 dB& PSNR\\
        Image&Input&AirNet~\cite{airnet}&PromptIR~\cite{promptir} &Ours&Reference\\
\end{tabular}
\end{center}
\caption{Image deraining comparisons on Rain100L~\cite{rain100L} under the three-degradation setting.}
\label{fig:aio1}
\end{figure*}

\begin{figure*}[!htb]
	\begin{center}
		\begin{tabular}{cccccc}

        \includegraphics[height=0.13\linewidth,width = 0.16\textwidth]{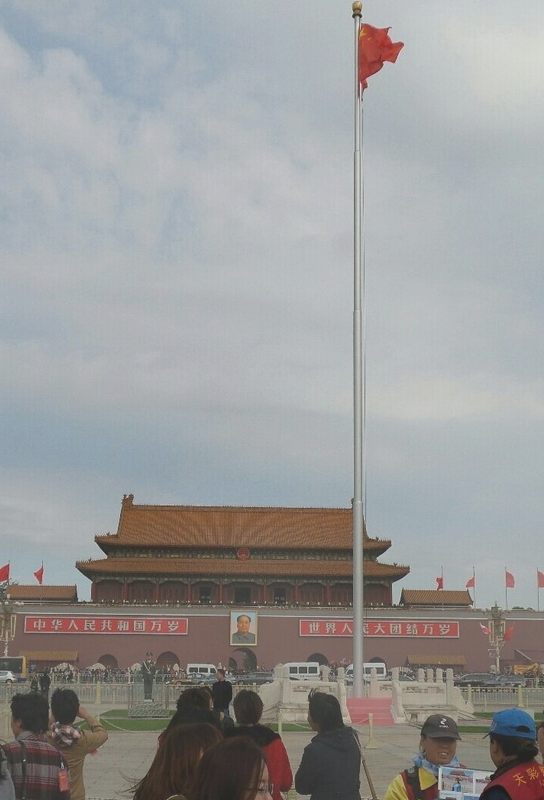}
        &\includegraphics[height=0.13\linewidth,width = 0.16\textwidth]{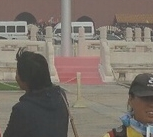}
        &\includegraphics[height=0.13\linewidth,width = 0.16\textwidth]{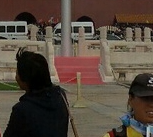}
        &\includegraphics[height=0.13\linewidth,width = 0.16\textwidth]{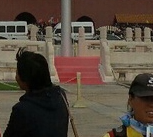} 
        &\includegraphics[height=0.13\linewidth,width = 0.16\textwidth]{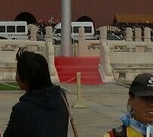} 
        &\includegraphics[height=0.13\linewidth,width = 0.16\textwidth]{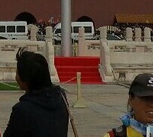} \\
        Degraded &14.65 dB&26.60 dB& 27.24 dB & 31.63 dB& PSNR\\

        \includegraphics[height=0.13\linewidth,width = 0.16\textwidth]{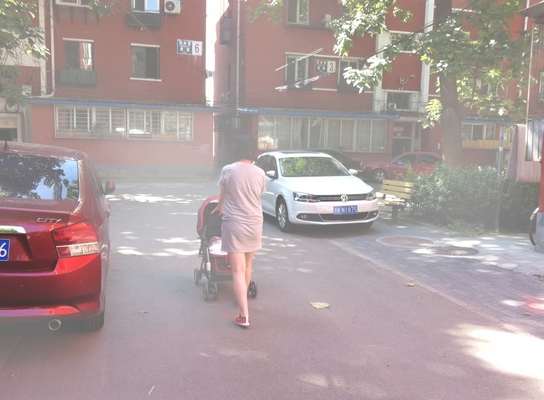}
        &\includegraphics[height=0.13\linewidth,width = 0.16\textwidth]{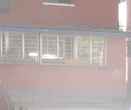}
        &\includegraphics[height=0.13\linewidth,width = 0.16\textwidth]{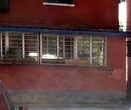}
        &\includegraphics[height=0.13\linewidth,width = 0.16\textwidth]{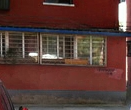} 
        &\includegraphics[height=0.13\linewidth,width = 0.16\textwidth]{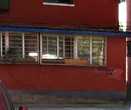} 
        &\includegraphics[height=0.13\linewidth,width = 0.16\textwidth]{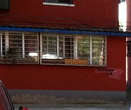} \\
        Degraded &6.58 dB&19.79 dB& 24.34 dB & 29.45 dB& PSNR\\

        \includegraphics[height=0.13\linewidth,width = 0.16\textwidth]{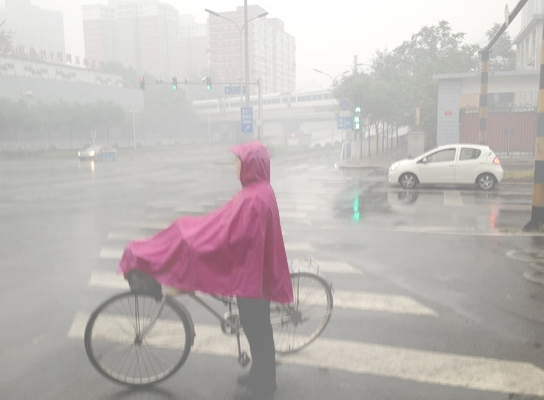}
        &\includegraphics[height=0.13\linewidth,width = 0.16\textwidth]{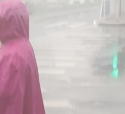}
        &\includegraphics[height=0.13\linewidth,width = 0.16\textwidth]{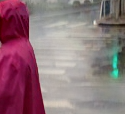}
        &\includegraphics[height=0.13\linewidth,width = 0.16\textwidth]{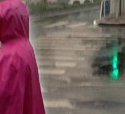} 
        &\includegraphics[height=0.13\linewidth,width = 0.16\textwidth]{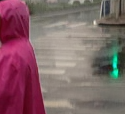} 
        &\includegraphics[height=0.13\linewidth,width = 0.16\textwidth]{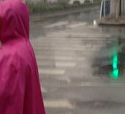} \\
        Degraded &9.61 dB&18.78 dB& 25.44 dB & 27.39 dB& PSNR\\

        \includegraphics[height=0.13\linewidth,width = 0.16\textwidth]{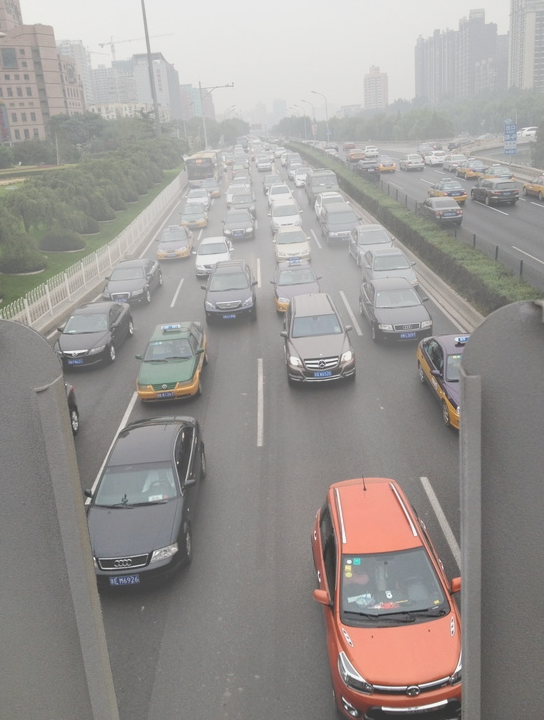}
        &\includegraphics[height=0.13\linewidth,width = 0.16\textwidth]{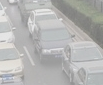}
        &\includegraphics[height=0.13\linewidth,width = 0.16\textwidth]{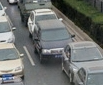}
        &\includegraphics[height=0.13\linewidth,width = 0.16\textwidth]{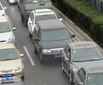} 
        &\includegraphics[height=0.13\linewidth,width = 0.16\textwidth]{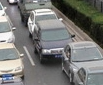} 
        &\includegraphics[height=0.13\linewidth,width = 0.16\textwidth]{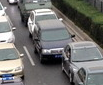} \\
        Degraded &10.49 dB&24.58 dB& 24.70 dB & 27.57 dB& PSNR\\        
        Image&Input&AirNet~\cite{airnet}&PromptIR~\cite{promptir} &Ours&Reference\\
\end{tabular}
\end{center}
\caption{Image dehazing comparisons on SOTS~\cite{RESIDE} under the three-degradation setting.}
\label{fig:aio2}
\end{figure*}

\vspace{-0.2cm}

\begin{figure*}[!htb]
	\begin{center}
\setlength{\abovecaptionskip}{-0.2cm}
		\begin{tabular}{cccccc}

        \includegraphics[height=0.13\linewidth,width = 0.16\textwidth]{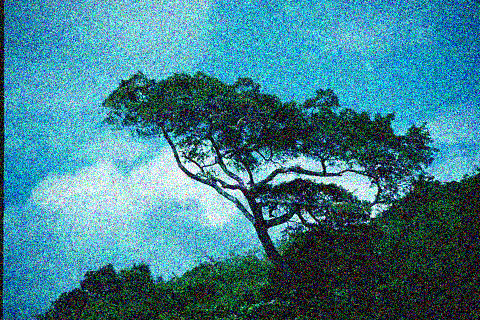}
        &\includegraphics[height=0.13\linewidth,width = 0.16\textwidth]{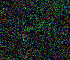}
        &\includegraphics[height=0.13\linewidth,width = 0.16\textwidth]{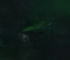}
        &\includegraphics[height=0.13\linewidth,width = 0.16\textwidth]{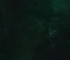} 
        &\includegraphics[height=0.13\linewidth,width = 0.16\textwidth]{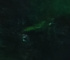} 
        &\includegraphics[height=0.13\linewidth,width = 0.16\textwidth]{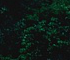} \\
        Degraded &16.49 dB& 28.73 dB& 28.77 dB & 29.09 dB& PSNR\\

        \includegraphics[height=0.13\linewidth,width = 0.16\textwidth]{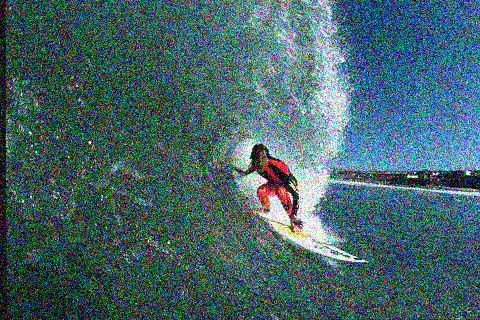}
        &\includegraphics[height=0.13\linewidth,width = 0.16\textwidth]{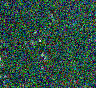}
        &\includegraphics[height=0.13\linewidth,width = 0.16\textwidth]{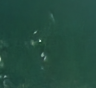}
        &\includegraphics[height=0.13\linewidth,width = 0.16\textwidth]{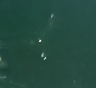} 
        &\includegraphics[height=0.13\linewidth,width = 0.16\textwidth]{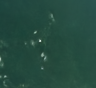} 
        &\includegraphics[height=0.13\linewidth,width = 0.16\textwidth]{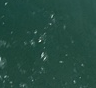} \\
        Degraded &15.13 dB& 31.68 dB& 31.50 dB &  31.99 dB& PSNR\\

        \includegraphics[height=0.13\linewidth,width = 0.16\textwidth]{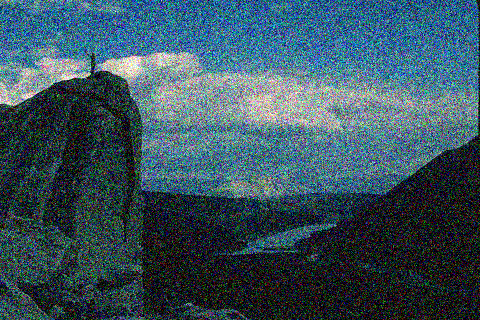}
        &\includegraphics[height=0.13\linewidth,width = 0.16\textwidth]{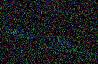}
        &\includegraphics[height=0.13\linewidth,width = 0.16\textwidth]{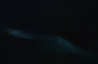}
        &\includegraphics[height=0.13\linewidth,width = 0.16\textwidth]{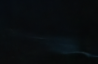} 
        &\includegraphics[height=0.13\linewidth,width = 0.16\textwidth]{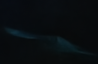} 
        &\includegraphics[height=0.13\linewidth,width = 0.16\textwidth]{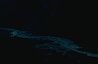} \\
        Degraded &16.90 dB& 36.02 dB& 34.08 dB &  36.31 dB& PSNR\\
        
        Image&Input&AirNet~\cite{airnet}&PromptIR~\cite{promptir} &Ours&Reference\\
\end{tabular}
\end{center}
\caption{Image denoising comparisons on BSD68~\cite{bsd68} with $\sigma=50$ under the three-degradation setting.}
\label{fig:aio3}
\end{figure*}

\begin{figure*}[!htb]
	\begin{center}
		\tabcolsep 1pt
		% [height=0.24\linewidth,width=0.24\linewidth]
% \resizebox{\linewidth}{!}{
		\begin{tabular}{cccc}

        \includegraphics[height=0.36\linewidth,width = 0.24\textwidth]{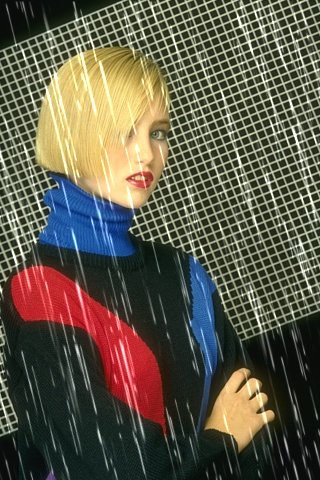}
        &\includegraphics[height=0.36\linewidth,width = 0.24\textwidth]{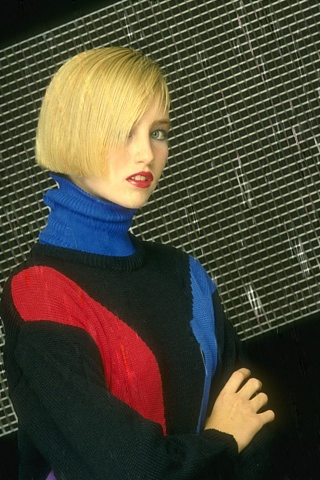}
        &\includegraphics[height=0.36\linewidth,width = 0.24\textwidth]{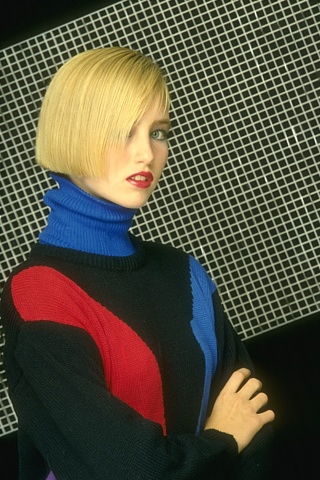}
        &\includegraphics[height=0.36\linewidth,width = 0.24\textwidth]{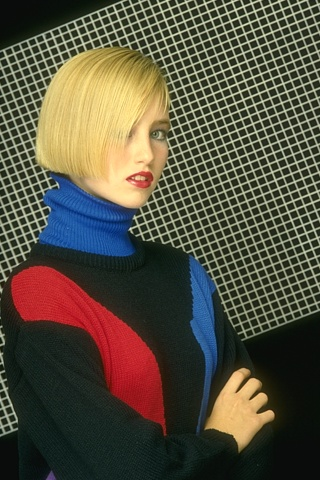} \\
        19.98 dB &18.85 dB&32.11 dB&PSNR\\       
        
        \includegraphics[height=0.36\linewidth,width = 0.24\textwidth]{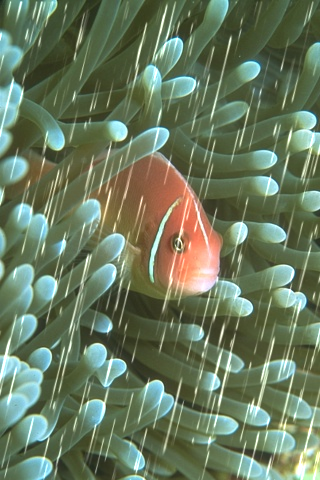}
        &\includegraphics[height=0.36\linewidth,width = 0.24\textwidth]{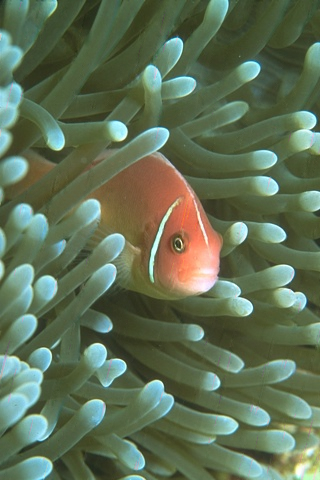}
        &\includegraphics[height=0.36\linewidth,width = 0.24\textwidth]{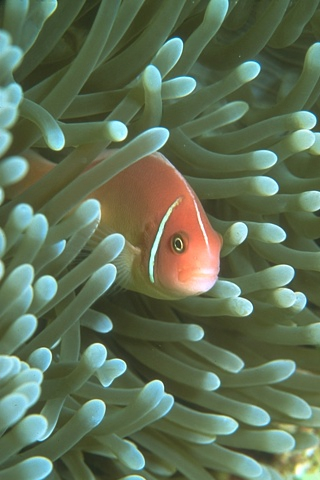}
        &\includegraphics[height=0.36\linewidth,width = 0.24\textwidth]{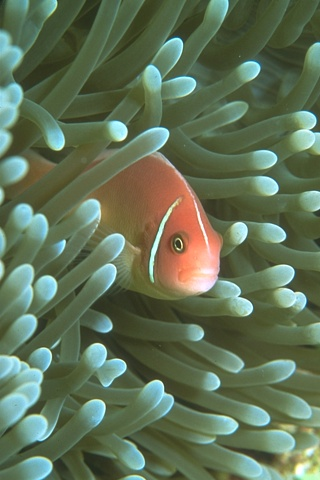} \\
        20.30 dB &35.08 dB&42.86 dB&PSNR\\

        \includegraphics[height=0.16\linewidth,width = 0.24\textwidth]{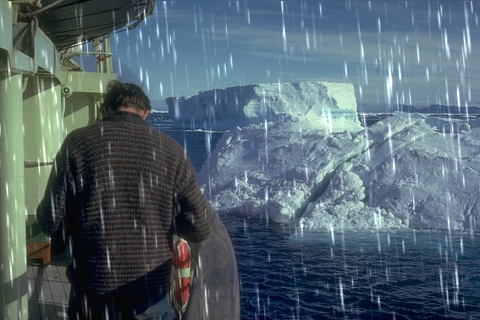}
        &\includegraphics[height=0.16\linewidth,width = 0.24\textwidth]{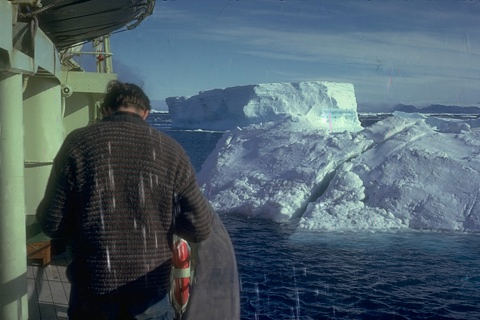}
        &\includegraphics[height=0.16\linewidth,width = 0.24\textwidth]{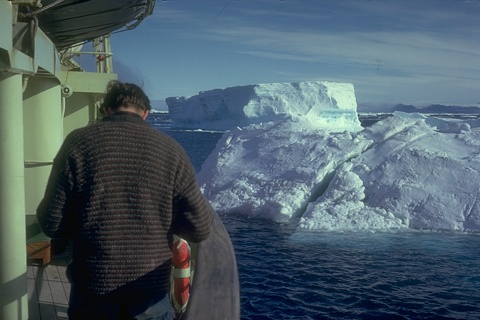}
        &\includegraphics[height=0.16\linewidth,width = 0.24\textwidth]{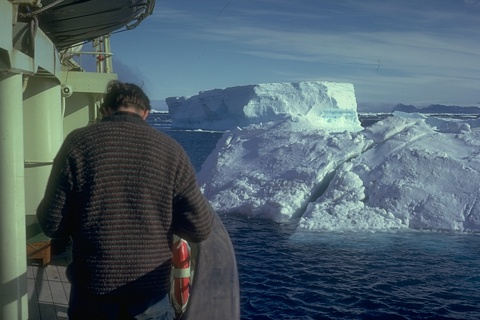} \\
        24.13 dB &33.50 dB&39.66 dB&PSNR\\
        
        \includegraphics[height=0.16\linewidth,width = 0.24\textwidth]{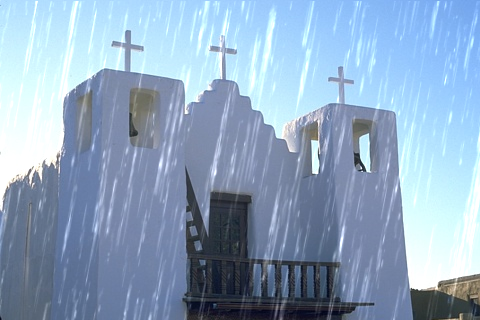}
        &\includegraphics[height=0.16\linewidth,width = 0.24\textwidth]{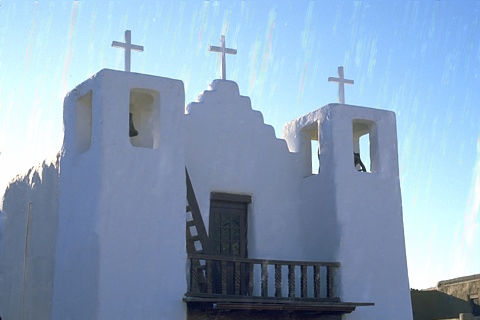}
        &\includegraphics[height=0.16\linewidth,width = 0.24\textwidth]{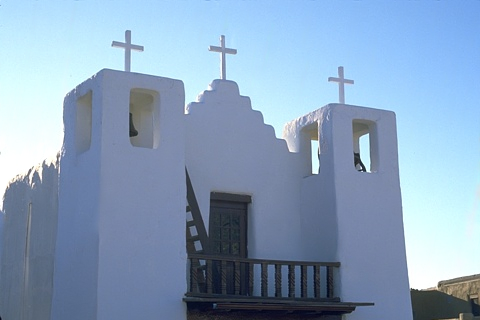}
        &\includegraphics[height=0.16\linewidth,width = 0.24\textwidth]{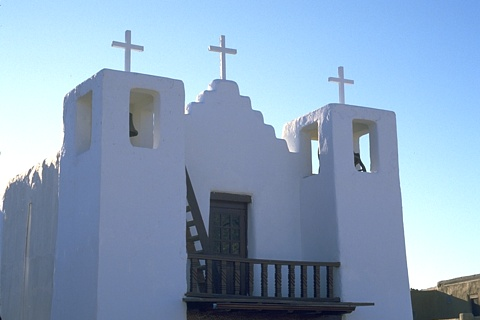} \\
        26.29 dB &35.39 dB&42.68 dB&PSNR\\

        \includegraphics[height=0.16\linewidth,width = 0.24\textwidth]{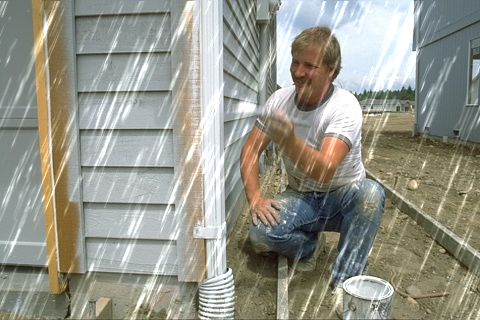}
        &\includegraphics[height=0.16\linewidth,width = 0.24\textwidth]{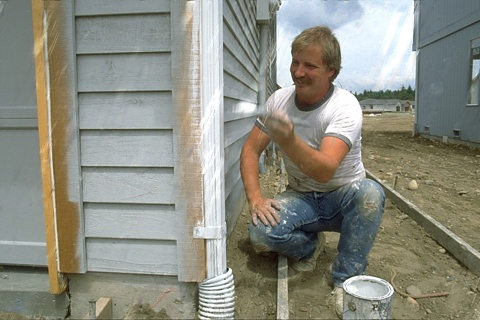}
        &\includegraphics[height=0.16\linewidth,width = 0.24\textwidth]{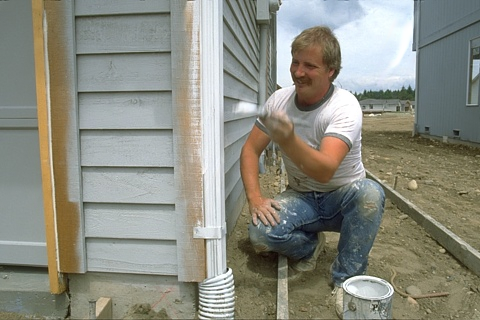}
        &\includegraphics[height=0.16\linewidth,width = 0.24\textwidth]{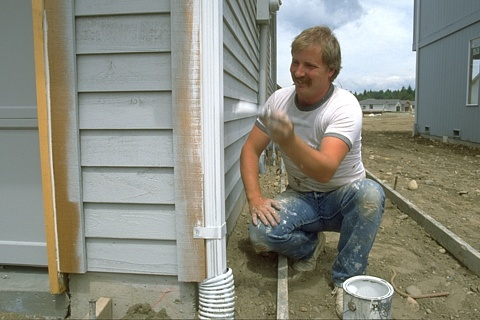} \\
        21.61 dB & 31.30 dB&35.57 dB&PSNR\\             
        Rainy Image &AirNet~\cite{airnet}&Ours&Reference\\
\end{tabular}
\end{center}
\caption{Image draining comparisons under the single task setting on Rain100L~\cite{rain100L}.}
\label{fig:deraining1}
\end{figure*}

\begin{figure*}[!htb]
	\begin{center}
		\tabcolsep 1pt
		% [height=0.24\linewidth,width=0.24\linewidth]
% \resizebox{\linewidth}{!}{
		\begin{tabular}{cccc}

        \includegraphics[height=0.18\linewidth,width = 0.24\textwidth]{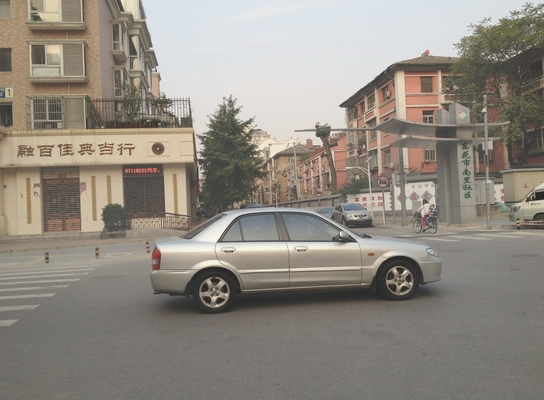}
        &\includegraphics[height=0.18\linewidth,width = 0.24\textwidth]{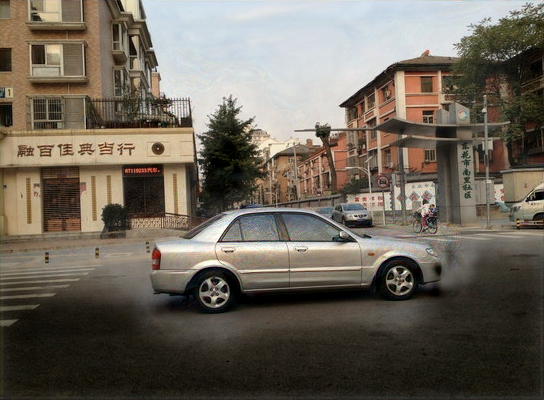}
        &\includegraphics[height=0.18\linewidth,width = 0.24\textwidth]{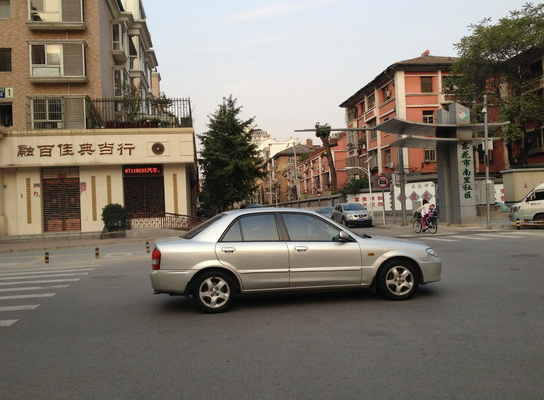}
        &\includegraphics[height=0.18\linewidth,width = 0.24\textwidth]{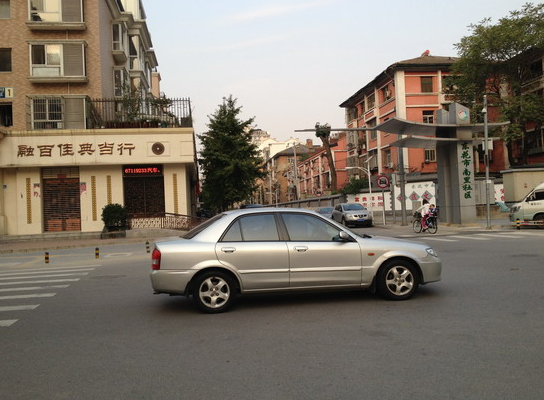} \\
        19.58 dB &17.81 dB&37.40 dB&PSNR\\

        \includegraphics[height=0.18\linewidth,width = 0.24\textwidth]{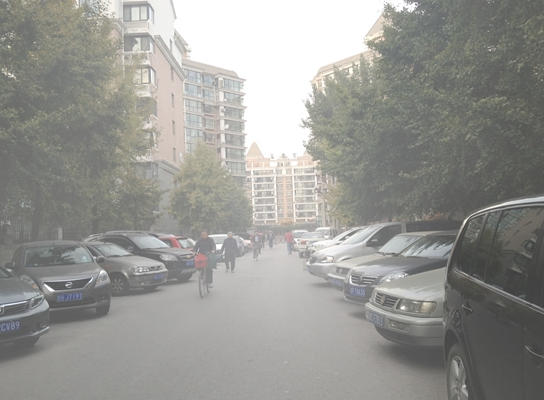}
        &\includegraphics[height=0.18\linewidth,width = 0.24\textwidth]{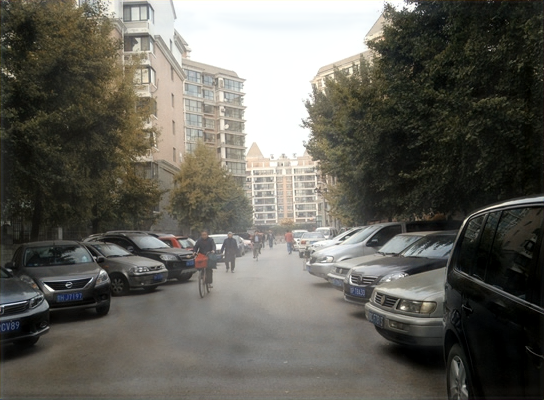}
        &\includegraphics[height=0.18\linewidth,width = 0.24\textwidth]{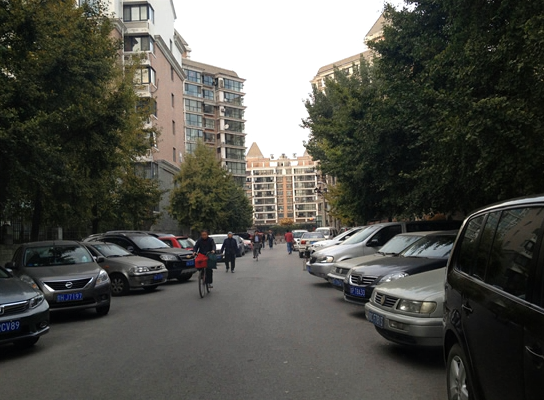}
        &\includegraphics[height=0.18\linewidth,width = 0.24\textwidth]{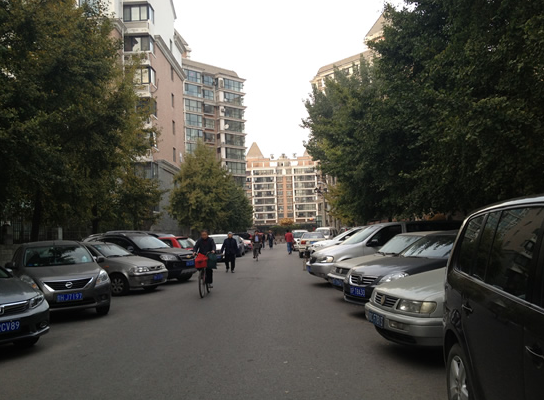} \\
        10.58 dB &20.12 dB&33.24 dB&PSNR\\

        \includegraphics[height=0.18\linewidth,width = 0.24\textwidth]{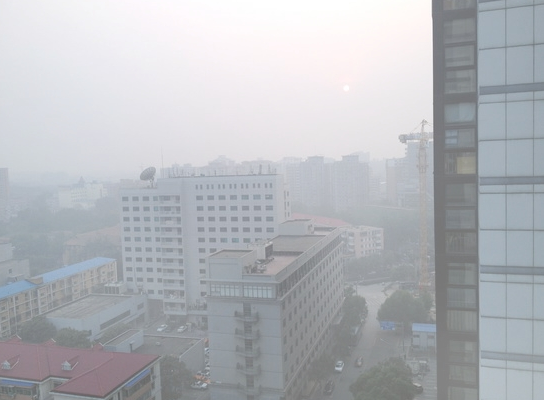}
        &\includegraphics[height=0.18\linewidth,width = 0.24\textwidth]{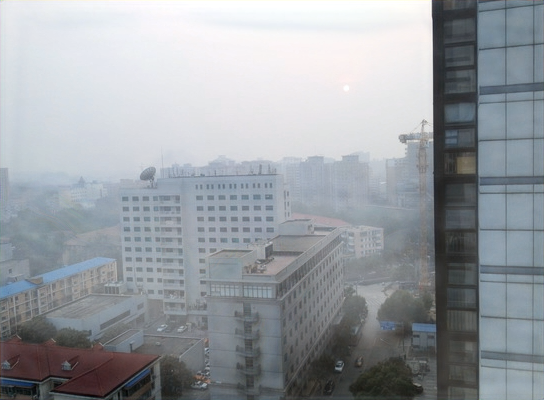}
        &\includegraphics[height=0.18\linewidth,width = 0.24\textwidth]{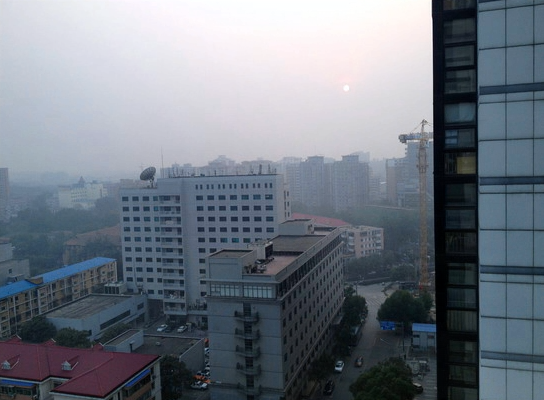}
        &\includegraphics[height=0.18\linewidth,width = 0.24\textwidth]{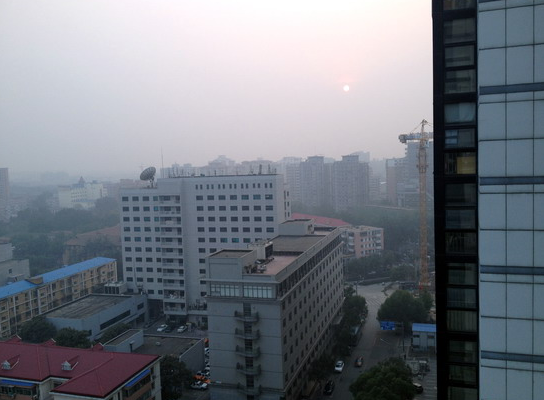} \\
        11.05 dB &15.59 dB&32.96 dB&PSNR\\    

        \includegraphics[height=0.18\linewidth,width = 0.24\textwidth]{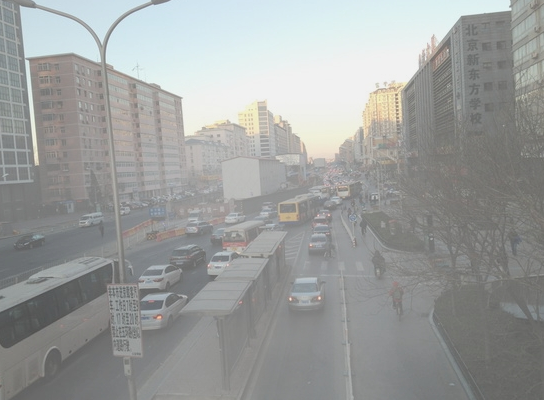}
        &\includegraphics[height=0.18\linewidth,width = 0.24\textwidth]{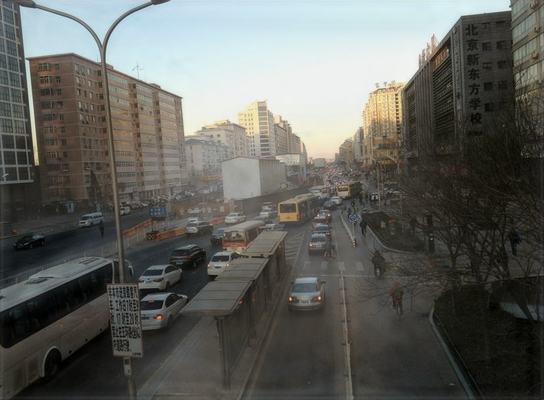}
        &\includegraphics[height=0.18\linewidth,width = 0.24\textwidth]{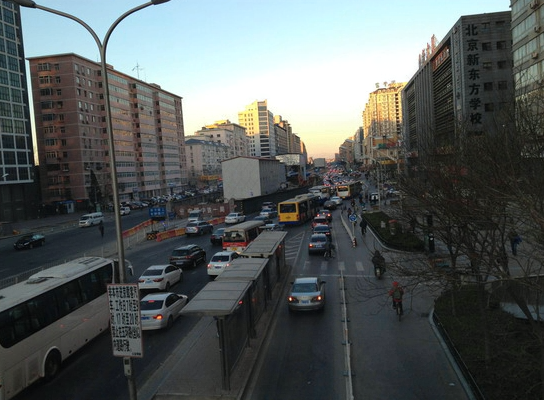}
        &\includegraphics[height=0.18\linewidth,width = 0.24\textwidth]{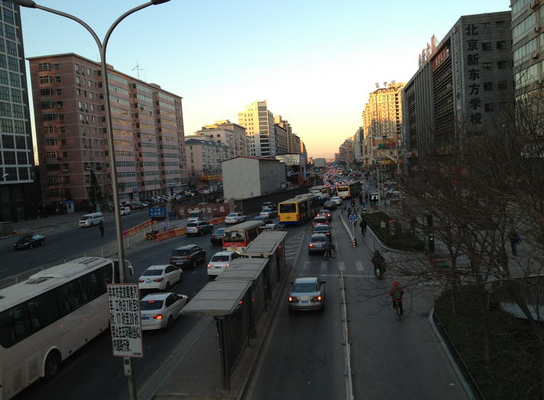} \\
        10.09 dB &21.28 dB&34.13 dB&PSNR\\

        \includegraphics[height=0.18\linewidth,width = 0.24\textwidth]{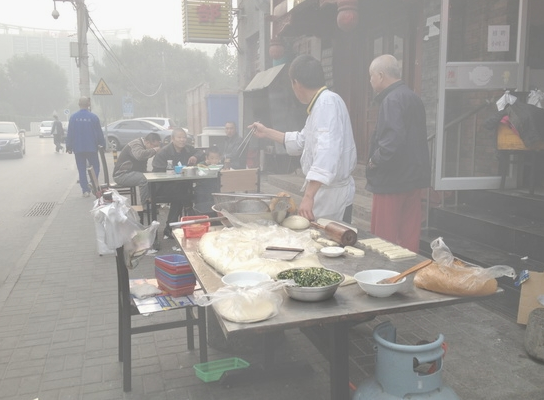}
        &\includegraphics[height=0.18\linewidth,width = 0.24\textwidth]{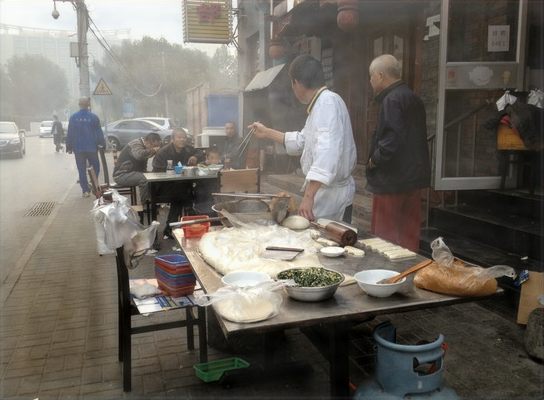}
        &\includegraphics[height=0.18\linewidth,width = 0.24\textwidth]{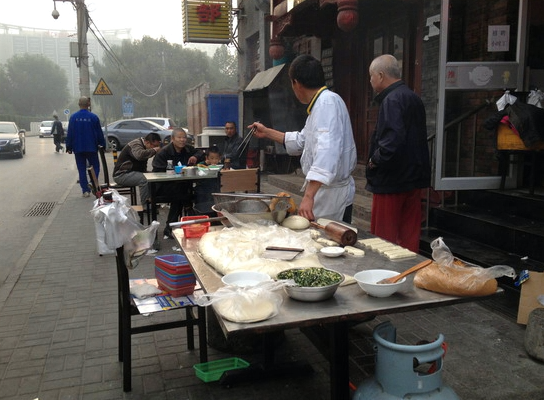}
        &\includegraphics[height=0.18\linewidth,width = 0.24\textwidth]{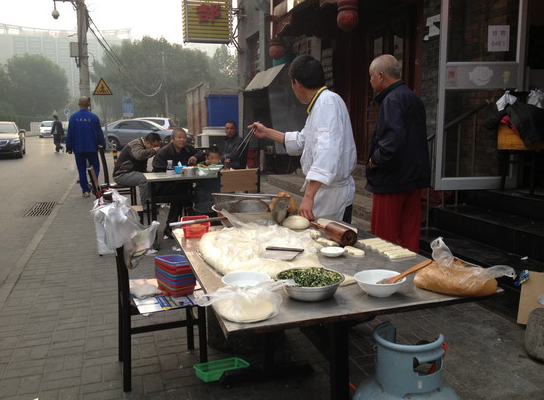} \\
        9.97 dB &17.61 dB&30.27 dB&PSNR\\

        \includegraphics[height=0.32\linewidth,width = 0.24\textwidth]{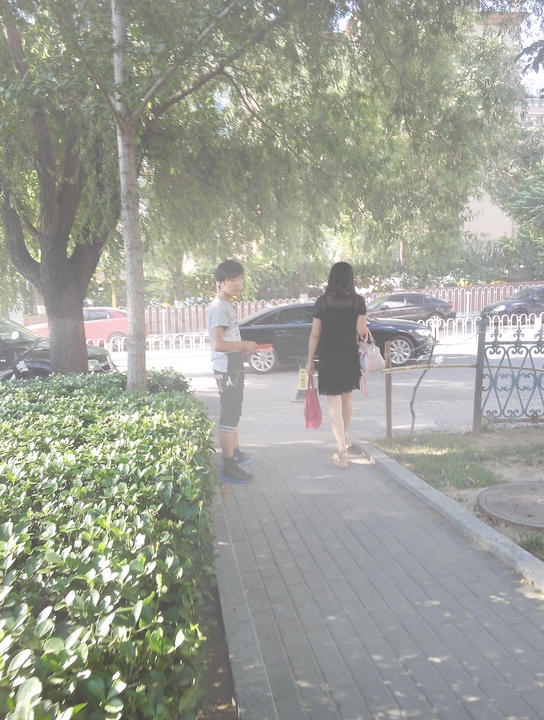}
        &\includegraphics[height=0.32\linewidth,width = 0.24\textwidth]{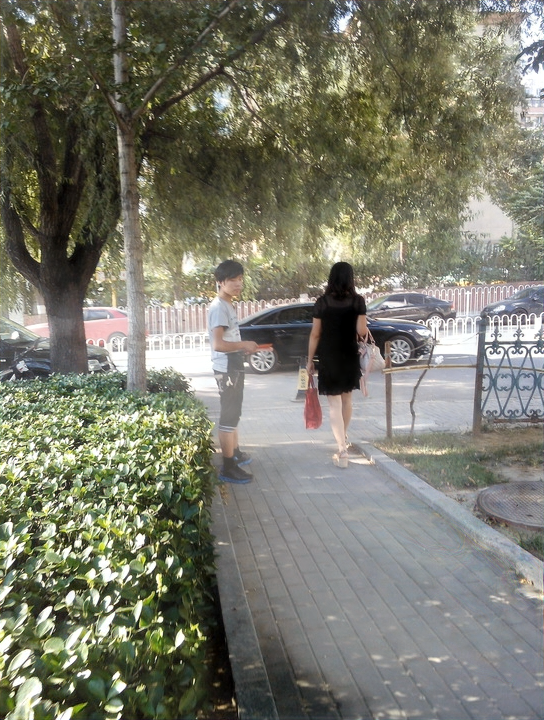}
        &\includegraphics[height=0.32\linewidth,width = 0.24\textwidth]{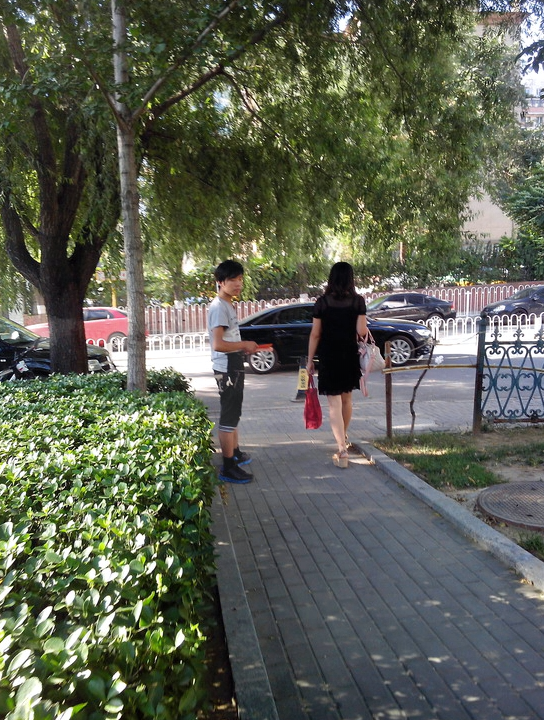}
        &\includegraphics[height=0.32\linewidth,width = 0.24\textwidth]{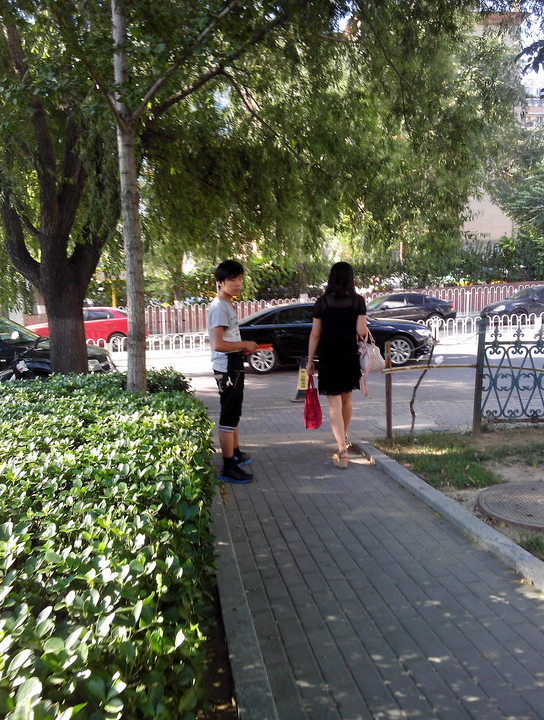} \\
        10.86 dB &16.98 dB&29.61 dB&PSNR\\

        Hazy Image &AirNet~\cite{airnet}&Ours&Reference\\
\end{tabular}
\end{center}
\caption{Image dehazing comparisons under the single task setting on SOTS~\cite{RESIDE}.}
\label{fig:dehazing1}
\end{figure*}

\begin{figure*}[!htb]
	\begin{center}
		\tabcolsep 1pt
		% [height=0.24\linewidth,width=0.24\linewidth]
% \resizebox{\linewidth}{!}{
		\begin{tabular}{ccccc}
        \includegraphics[height=0.14\linewidth,width = 0.19\textwidth]{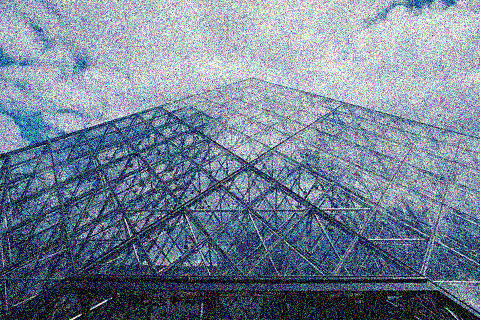}     
        &\includegraphics[height=0.14\linewidth,width = 0.19\textwidth]{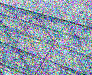}
        &\includegraphics[height=0.14\linewidth,width = 0.19\textwidth]{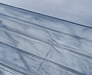}
        &\includegraphics[height=0.14\linewidth,width = 0.19\textwidth]{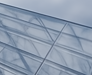}
        &\includegraphics[height=0.14\linewidth,width = 0.19\textwidth]{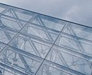} \\
        Noisy&14.49 dB &27.49 dB&28.69 dB&PSNR\\
        
        \includegraphics[height=0.14\linewidth,width = 0.19\textwidth]{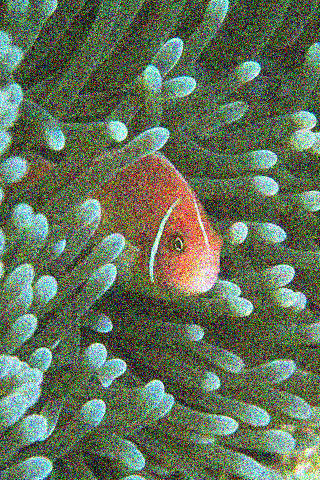} 
        &\includegraphics[height=0.14\linewidth,width = 0.19\textwidth]{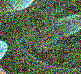} &
        \includegraphics[height=0.14\linewidth,width = 0.19\textwidth]{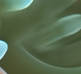}
        &\includegraphics[height=0.14\linewidth,width = 0.19\textwidth]{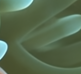}
        &\includegraphics[height=0.14\linewidth,width = 0.19\textwidth]{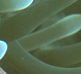}\\
        Noisy&14.58 dB &31.97 dB&32.57 dB& PSNR\\

        \includegraphics[height=0.14\linewidth,width = 0.19\textwidth]{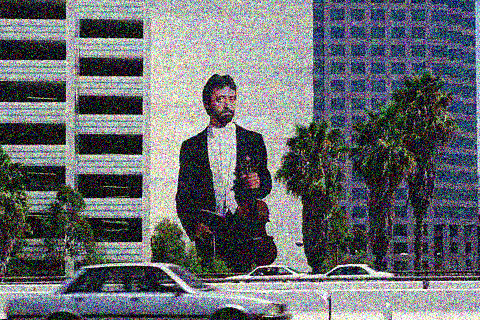}
        &\includegraphics[height=0.14\linewidth,width = 0.19\textwidth]{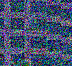}
        &\includegraphics[height=0.14\linewidth,width = 0.19\textwidth]{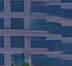}
        &\includegraphics[height=0.14\linewidth,width = 0.19\textwidth]{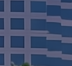}
        &\includegraphics[height=0.14\linewidth,width = 0.19\textwidth]{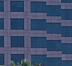} \\
        Noisy&14.69 dB &29.18 dB&31.17 dB&PSNR\\

        \includegraphics[height=0.14\linewidth,width = 0.19\textwidth]{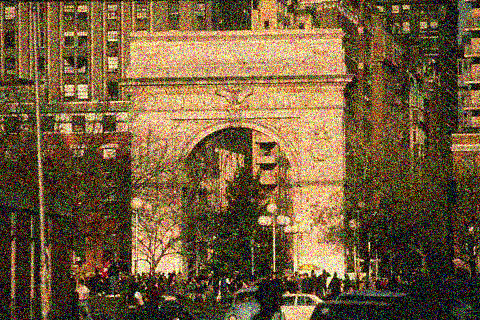}
        &\includegraphics[height=0.14\linewidth,width = 0.19\textwidth]{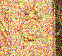}
        &\includegraphics[height=0.14\linewidth,width = 0.19\textwidth]{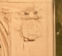}
        &\includegraphics[height=0.14\linewidth,width = 0.19\textwidth]{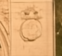}
        &\includegraphics[height=0.14\linewidth,width = 0.19\textwidth]{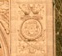} \\
        Noisy&15.08 dB &26.73 dB&27.31 dB& PSNR\\

        \includegraphics[height=0.14\linewidth,width = 0.19\textwidth]{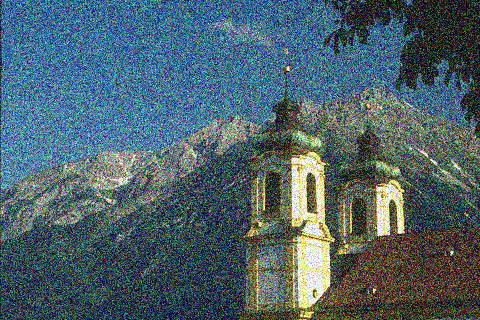}
        &\includegraphics[height=0.14\linewidth,width = 0.19\textwidth]{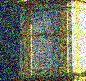}
        &\includegraphics[height=0.14\linewidth,width = 0.19\textwidth]{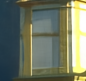}
        &\includegraphics[height=0.14\linewidth,width = 0.19\textwidth]{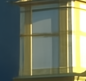}
        &\includegraphics[height=0.14\linewidth,width = 0.19\textwidth]{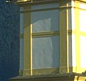} \\
        Noisy&14.87 dB &29.25 dB&29.75 dB& PSNR\\

        \includegraphics[height=0.14\linewidth,width = 0.19\textwidth]{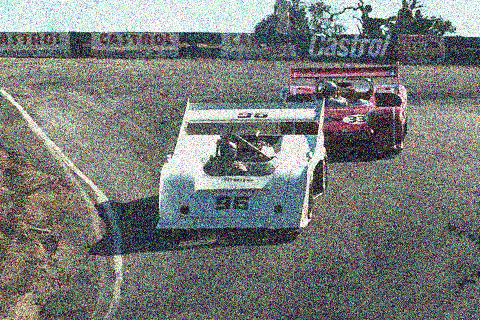}
        &\includegraphics[height=0.14\linewidth,width = 0.19\textwidth]{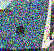}
        &\includegraphics[height=0.14\linewidth,width = 0.19\textwidth]{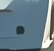}
        &\includegraphics[height=0.14\linewidth,width = 0.19\textwidth]{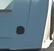}
        &\includegraphics[height=0.14\linewidth,width = 0.19\textwidth]{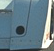} \\
        Noisy&  14.62 dB & 29.74 dB&30.44 dB& PSNR\\
          
        \includegraphics[height=0.14\linewidth,width = 0.19\textwidth]{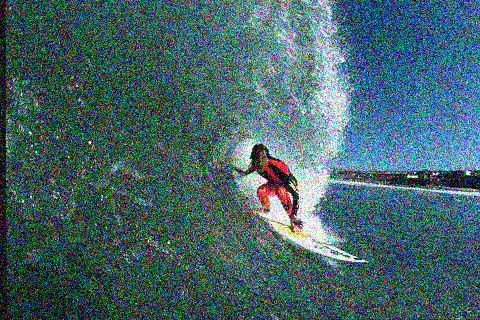}
        &\includegraphics[height=0.14\linewidth,width = 0.19\textwidth]{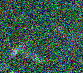}
        &\includegraphics[height=0.14\linewidth,width = 0.19\textwidth]{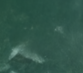}
        &\includegraphics[height=0.14\linewidth,width = 0.19\textwidth]{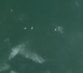}
        &\includegraphics[height=0.14\linewidth,width = 0.19\textwidth]{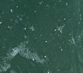} \\
        Noisy&14.83 dB & 29.87 dB& 30.19 dB&PSNR\\      
        Image &Input&AirNet~\cite{airnet}&Ours&Reference\\
\end{tabular}
\end{center}
\caption{Image denoising results under single task setting on BSD68~\cite{bsd68} with $\sigma=50$.}
\label{fig:denoising2}
\end{figure*}

\end{document}